\documentclass[runningheads]{llncs}

\usepackage{eccv}

\usepackage{eccvabbrv}

\usepackage{graphicx}
\usepackage{booktabs}

\usepackage[accsupp]{axessibility}  

\usepackage[pagebackref,breaklinks,colorlinks,citecolor=eccvblue]{hyperref}

\definecolor{vae}{HTML}{3b50af} 
\definecolor{initial_surface}{HTML}{6c39d2} 
\definecolor{generation}{HTML}{ab4202} 

\definecolor{cvprblue}{rgb}{0.21,0.49,0.74}
\usepackage{graphicx}
\usepackage{multirow}
\usepackage{colortbl}
\usepackage{rotating}
\usepackage{setspace}
\usepackage{booktabs}
\usepackage{float}
\usepackage{amssymb}
\usepackage{pifont}
\usepackage{csquotes}
\usepackage{algorithm}
\usepackage{algpseudocode}
\usepackage{tikz}
\usetikzlibrary{calc}
\usepackage{amsmath}
\usepackage{tabularray}
\usepackage[table, dvipsnames, xcdraw]{xcolor}
\usepackage{lipsum}
\usepackage{arydshln}

\usepackage{orcidlink}

\definecolor{codebg}{HTML}{F7F7F7}   
\definecolor{kw}{HTML}{1A4E9A}       
\definecolor{cm}{HTML}{888888}       
\definecolor{op}{HTML}{D95F02}       

\usepackage{minted}
\setminted{
    fontsize=\small,
    bgcolor=codebg,
    baselinestretch=1.15,
    breaklines=true,
    style=default,
    python3=true
}

\definecolor{nearblack}{rgb}{0.15,0.15,0.15}
\definecolor{charcoal}{rgb}{0.20,0.20,0.20}
\definecolor{deepnavy}{rgb}{0.05,0.20,0.40}
\definecolor{slateblue}{rgb}{0.10,0.18,0.35}
\definecolor{steelindigo}{rgb}{0.15,0.25,0.45}
\definecolor{darkindigo}{rgb}{0.25,0.20,0.45}
\definecolor{deepplum}{rgb}{0.20,0.15,0.30}
\definecolor{aubergine}{rgb}{0.25,0.20,0.25}
\definecolor{royalpurple}{rgb}{0.30,0.15,0.35}
\definecolor{espressobrown}{rgb}{0.35,0.20,0.15}
\definecolor{darkumber}{rgb}{0.45,0.30,0.20}
\definecolor{olivedrab}{rgb}{0.20,0.25,0.15}
\definecolor{pinegreen}{rgb}{0.10,0.25,0.20}
\definecolor{junglegreen}{rgb}{0.15,0.30,0.25}
\definecolor{stormblue}{rgb}{0.15,0.20,0.30}
\definecolor{deepteal}{rgb}{0.20,0.28,0.35}
\definecolor{mossgreen}{rgb}{0.25,0.30,0.20}
\definecolor{mulberry}{rgb}{0.30,0.20,0.25}
\definecolor{crimsonwine}{rgb}{0.35,0.15,0.25}
\definecolor{raspberryblack}{rgb}{0.28,0.12,0.22}
\definecolor{twilightviolet}{rgb}{0.22,0.18,0.32}
\definecolor{petrolblue}{rgb}{0.18,0.24,0.28}
\definecolor{earthykhaki}{rgb}{0.28,0.24,0.12}
\definecolor{stonebrown}{rgb}{0.32,0.28,0.20}
\definecolor{duskpurple}{rgb}{0.22,0.20,0.28}
\definecolor{rosewood}{rgb}{0.34,0.22,0.30}
\definecolor{huntergreen}{rgb}{0.18,0.22,0.20}
\definecolor{darkslate}{rgb}{0.12,0.18,0.22}
\definecolor{obsidianteal}{rgb}{0.08,0.16,0.18}
\definecolor{midnightblue}{rgb}{0.25,0.25,0.40}
\definecolor{oxblood}{rgb}{0.40,0.15,0.10}
\definecolor{bloodred}{rgb}{0.26,0.14,0.18}
\definecolor{brickred}{rgb}{0.28,0.18,0.20}
\definecolor{dustycrimson}{rgb}{0.32,0.18,0.22}
\definecolor{maroonwine}{rgb}{0.25,0.12,0.18}

\newcommand{\gchar}[2]{{\color[HTML]{#1}#2}\discretionary{}{}{}}
\newcommand{\gradienturl}{%
  \href{https://sadilkhan.github.io/dreamcad2026/}{%
    \gchar{2563EB}{h}\gchar{2B5EE9}{t}\gchar{3059E7}{t}\gchar{3554E5}{p}%
    \gchar{3A4FE3}{s}\gchar{3F4AE1}{:}\gchar{4445DF}{/}\gchar{4940DD}{/}%
    \gchar{4E3BDB}{s}\gchar{5336D9}{a}\gchar{5831D7}{d}\gchar{5D2CD5}{i}%
    \gchar{6227D3}{l}\gchar{6722D1}{k}\gchar{6C1DCF}{h}\gchar{7118CD}{a}%
    \gchar{7613CB}{n}\gchar{7B0EC9}{.}\gchar{8009C7}{g}\gchar{8504C5}{i}%
    \gchar{8A00C3}{t}\gchar{8F00BE}{h}\gchar{9400B9}{u}\gchar{9900B4}{b}%
    \gchar{9E00AF}{.}\gchar{A300AA}{i}\gchar{A800A5}{o}\gchar{AD00A0}{/}%
    \gchar{B2009B}{d}\gchar{B70096}{r}\gchar{BC0091}{e}\gchar{C1008C}{a}%
    \gchar{C60087}{m}\gchar{CB0082}{c}\gchar{D0007D}{a}\gchar{D50078}{d}%
    \gchar{D50060}{2}\gchar{D50048}{0}\gchar{D50030}{2}\gchar{D50018}{6}%
    \gchar{D50000}{/}%
  }%
}

\newcommand{\colname}[2]{\textcolor{#1}{\textbf{#2}}}

\definecolor{algstep}{RGB}{30,70,160}      
\definecolor{algvar}{RGB}{180,60,60}       
\definecolor{alghigh}{RGB}{0,120,90}       

\definecolor{text2cad}{HTML}{066236}
\definecolor{point2cad}{HTML}{bd5616}
\definecolor{image2cad}{HTML}{5271ff}
\definecolor{ubc}{HTML}{048A81}
\usepackage{pifont}
\usepackage{graphicx}
\newcommand{\cmark}{\textcolor{green!60!black}{\ding{51}}} 
\newcommand{\xmark}{\textcolor{red!70!black}{\ding{55}}}   

\begin{document}


\title{%
\texorpdfstring{%
\raisebox{-0.05cm}{\includegraphics[height=0.5cm]{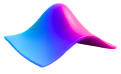}}\hspace{0.1cm}\raisebox{-0.03cm}{\includegraphics[height=0.5cm]{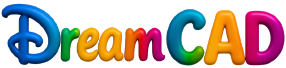}}%
\raisebox{0.1em}{:} \textbf{Scaling Multi-modal CAD Generation using Differentiable Parametric Surfaces}%
}{%
DreamCAD: Scaling Multi-modal CAD Generation using Differentiable Parametric Surfaces%
}%
}

\titlerunning{DreamCAD: Scaling Multi-modal CAD Generation}

\author{Muhammad Sadil Khan\inst{1,2}\thanks{Work done during an internship at Huawei London Research Center.} \and
Muhammad Usama\inst{1,2} \and
Rolandos Alexandros Potamias\inst{3} \and
Didier Stricker\inst{1,2} \and
Muhammad Zeshan Afzal\inst{1} \and
Jiankang Deng\inst{3\dagger} \and
Ismail Elezi\inst{4}}

\authorrunning{M.S. Khan et al.}

\institute{DFKI, Germany \and
RPTU Kaiserslautern, Germany \and
Imperial College London, United Kingdom \and
Huawei London Research Center, United Kingdom}

\maketitle

\begingroup
\renewcommand\thefootnote{\dag}
\footnotetext{Corresponding author.}
\endgroup


\begin{abstract}
Multimodal CAD generation faces a fundamental scalability 
challenge. Design-history methods are confined to small 
annotated datasets, while BRep topology is discrete and 
non-differentiable. Meanwhile, millions of unannotated 3D 
meshes remain untapped, since existing CAD methods cannot 
leverage them without explicit CAD 
annotations. We propose \textbf{DreamCAD}, a multimodal 
generative framework that bridges this gap by representing 
shapes as $C^0$-continuous Bézier patches with 
differentiable tessellation, enabling direct point-level 
supervision on large-scale 3D meshes without CAD-specific 
annotations. We further 
introduce \textbf{CADCap-1M}, the largest CAD captioning 
dataset with $1$M+ GPT-5-generated descriptions to advance 
text-to-CAD research. DreamCAD achieves state-of-the-art 
performance on ABC and Objaverse across text, image, and 
point modalities, surpassing 75\% user preference. 
Finally, we show that DreamCAD's accurate, compact geometry enables topology recovery into production-ready CAD models, exportable as editable STEP files. Project page is available at \gradienturl{}.

    
\end{abstract}
\section{Introduction}

Computer-Aided Design forms the foundation of modern 
engineering~\cite{zou2024intelligentcad20}, architecture, 
and manufacturing~\cite{application-cad}. Unlike 3D meshes 
or point clouds that approximate geometry~\cite{3d-survey-1}, 
CAD models are built from parametric primitives such as 
Bézier and NURBS surfaces, encoded in boundary representation 
(BRep) format, capturing both topology and precise geometry 
essential for industrial design~\cite{param-complexity}. 
With the advent of generative AI~\cite{stable-diff3,trellis}, 
AI-assisted CAD generation~\cite{sharp2023} has become a 
promising frontier for accelerating design workflows and 
enabling rapid prototyping of manufacturable 
models~\cite{future-cad}.

\begin{figure}[t]
    \centering
    \includegraphics[width=0.95\linewidth]{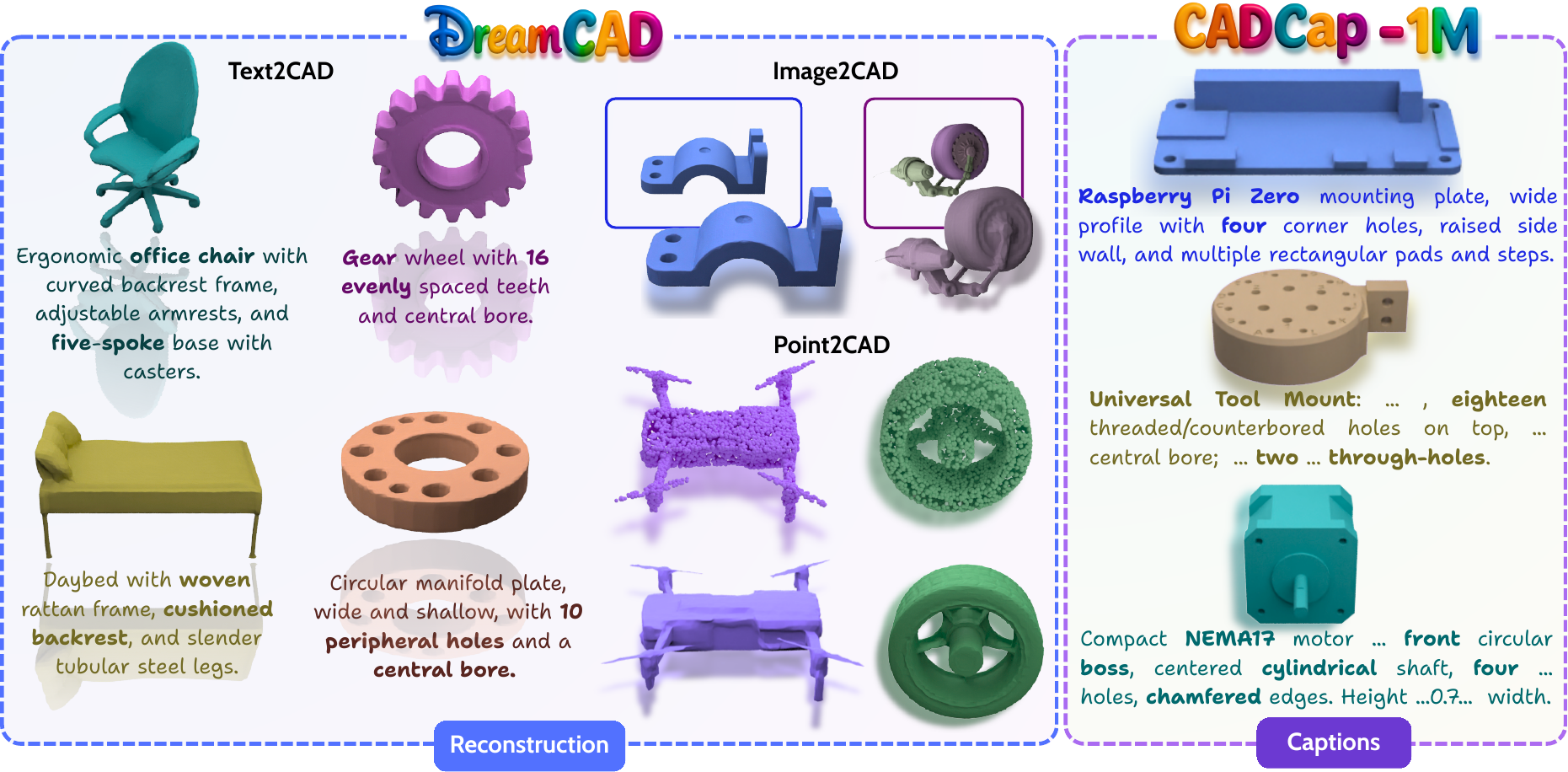}
    \caption{Our proposed \textbf{DreamCAD} (left) is a multimodal generative framework that can reconstruct CAD models from text, images, and point clouds using parametric patches. \textbf{CADCap-1M} (right) provides 1M+ GPT-5–generated captions.}
    \label{fig:teaser}
\end{figure}

\vspace{0.1cm}
\noindent Recent CAD generation methods enable multimodal synthesis from text~\cite{text2cad}, images~\cite{img2cad}, or point clouds~\cite{point2cad}. However, achieving strong generalization across modalities and diverse geometries remains a major challenge. The root cause lies in the CAD representation and the training scheme. Design-history-based models~\cite{deepcad} rely on \textit{sketch-and-extrude} sequences from small datasets such as DeepCAD-160k~\cite{deepcad} and Fusion360-8k~\cite{fusion360}, constraining the generalization in freeform or open-vocabulary shapes. UV and graph-based methods~\cite{brepgen} require explicit BRep annotations, which are costly and difficult to scale. Although ABC~\cite{abc} contains 1M BReps, it remains underutilized due to these bottlenecks. 

 \vspace{0.1cm}
\noindent In contrast, modern 3D generative models~\cite{hunyuan3d} scale effectively by decoupling mesh generation into multiple stages, using intermediate representations such as SDFs \cite{sdf}, or NeRFs~\cite{nerfs}. We argue that scaling multimodal CAD generation requires an analogous paradigm shift - from single-stage joint geometry and topology generation to a decoupled pipeline. The first stage should focus on scalable parametric geometry leveraging unstructured 3D data without explicit CAD annotations, while subsequent stages can work toward recovering full CAD topology from this intermediate geometric foundation.

 \vspace{0.1cm}
 \noindent To address the first stage of generalizability, we propose \textbf{DreamCAD}, a multi-modal generative framework that produces editable parametric surfaces from point-level supervision alone. Each CAD model is represented as a set of rational Bézier patches defined by learnable control points and weights. These surfaces are differentiably tessellated into meshes for point-based supervision via Chamfer loss. DreamCAD encodes sparse voxels into structured latents (SLAT)~\cite{trellis} to learn a CAD-oriented latent space, which is decoded into parametric geometry. Building on this, we adopt a coarse-to-fine conditional generation framework supporting \textit{text}, \textit{image}, and \textit{point cloud} inputs.

\vspace{0.1cm}
\noindent A key challenge with patch-based BRep representations is ensuring $C^0$ continuity, which is essential for valid CAD modeling. $C^0$ continuity~\cite{C0} requires adjacent surface patches to share common boundaries without gaps or overlaps. We address the continuity problem structurally rather than using any geometric optimization. Starting from a sparse voxel grid, we remove internal quads via flood-fill~\cite{flood-fill}. This way, each surface quad maps to a parametric patch initialized with control-point grids and unit weights, with adjacent patches sharing boundary points for continuity. The VAE decoder then refines control points and weights to match the target geometry. As a downstream application, in 
Section~\ref{sec:topology}, we show that DreamCAD's high-fidelity and compact geometric 
reconstruction can provide a strong geometric prior for topology recovery.


\begin{table}[t]
\def\arraystretch{1.2}%
\caption{Training datasets for DreamCAD comprising 1.3M 3D meshes. \cmark denotes if the dataset is included in proposed CADCap-1M dataset.}
\label{tab:dreamcad_training_dataset}
\centering
\setlength{\tabcolsep}{6pt}
\resizebox{0.75\columnwidth}{!}{
\begin{tabular}{lcc:lcc}
\hline
Dataset& Samples & CADCap\quad & Dataset& Samples & CADCap\\ \hline
ABC~\cite{abc} & 757,433 & \cmark & 3D-Future~\cite{3d-future} & 16,990 & \cmark\\
Automate~\cite{automate} & 380,124 & \cmark & ModelNet~\cite{modelnet} & 12,308 & \cmark\\
ShapeNet~\cite{shapenet} & 52,458 & \xmark &  ABO~\cite{abo} & 7,944 & \xmark\\
CADParser~\cite{cadparser} & 40,989 & \cmark & Fusion360~\cite{fusion360} & 4,603 & \cmark\\
HSSD~\cite{hssd} & 30,078 & \xmark & Toys4K~\cite{toys4k} & 3,482 & \xmark\\
\hline
\rowcolor{blue!10}
\multicolumn{2}{l}{\textbf{Total}} & & & \textbf{1,306,409} & \\
\hline
\end{tabular}    
}
\end{table}

\vspace{0.1cm}
\noindent We curate over 1M 3D meshes from 10 publicly available datasets, as summarized in Table~\ref{tab:dreamcad_training_dataset}, to train our VAE architecture. For \textit{text-to-CAD} training, existing datasets such as Text2CAD-$160$K~\cite{text2cad} lack the scale and diversity necessary for robust generative modeling. To address this, we construct \textbf{CADCap-1M}, the largest CAD captioning dataset to date. We leverage GPT-5 to generate high-quality captions for $1$M+ models from existing large-scale CAD datasets. We evaluate DreamCAD on multimodal generation tasks across \textit{text}, \textit{image}, and \textit{point cloud} on both ABC~\cite{abc} and Objaverse~\cite{objaverse} datasets. Our experiments show that DreamCAD consistently surpasses baselines in both geometric accuracy and user preference. Our contributions are as follows
\vspace*{-.3\baselineskip}
\begin{enumerate}
    \item We introduce \textbf{DreamCAD}, a multi-modal generative framework trained only with point supervision without any dependence on CAD annotations.
    \item We release \textbf{CADCap-1M}, the largest CAD captioning dataset with over $1$M text descriptions for scalable text-to-CAD research.
    \item DreamCAD achieves state-of-the-art performance across \textit{point}-, \textit{image}-, and \textit{text}-conditioned generation tasks, reducing Chamfer Distance by up to 70\% in \textit{point-to-CAD} and surpassing 75\% preference in expert and GPT-based evaluations for \textit{text} and \textit{image-to-CAD}.

\end{enumerate}
\section{Related Work}

\noindent \textbf{Generative CAD.} Prior CAD generative 
models~\cite{deepcad,brepnet,nurbgen} represent geometry 
through boundary representations (BReps), which offer 
exact, watertight, and editable geometry but are 
difficult to learn at scale due to their complex 
parametric and topological structure~\cite{complexgen}. Existing approaches therefore struggle to balance scalability with geometric fidelity.

\vspace{0.1cm}
\noindent Design-history methods~\cite{deepcad,text2cad,cadrecode} model CAD creation using \textit{sketch-extrude} operations, framed as program synthesis~\cite{cadparser,cadrecode,cadgpt,text2cadquery} or language modeling~\cite{cadllama,cad-instruct,cadsignet,transcad}. However, they require design-history logs available only in proprietary CAD softwares~\cite{autodesk,onshape}. This limits training to small datasets (e.g., DeepCAD-170K~\cite{deepcad}, Fusion360-8K~\cite{fusion360}) and models exhibit poor generalization to complex, real-world geometries~\cite{cadmium,cadgpt}. 

\vspace{0.1cm}
\noindent Some works model BReps through UV parameterizations or hierarchical graph structures since explicit BRep topology (\eg vertices, edges, faces) is discrete and non-differentiable, preventing gradient-based optimization for geometry based learning. UVNet~\cite{uvnet} and DTG-BRepGen~\cite{dtgbrepgen} predict UV grids from face annotations, while BRepGen~\cite{brepgen} and BRepDiff~\cite{brepdiff} perform UV-space diffusion with post-processing. However, these methods face some key limitations: (i) no guaranteed $C^0$ continuity across adjacent faces, (ii) training restricted to models with fewer faces~\cite{brepdiff}, excluding $\sim$70\% of ABC~\cite{abc}, (iii) BRep conversion requires expensive grid-based fitting stages with high error and invaldity rate~\cite{brepdetnet,complexgen}, and (iv) UV parameterizations only approximate geometry with resolution-related computational challenges. Self-supervised methods~\cite{secadnet,extrudenet,point2cyl} attempts to learn CAD geometry without using any CAD annotation. However, these methods either do not scale well or restrict to simpler shapes and have never been explored for multimodal purposes. Similarly, surface-fitting approaches~\cite{point2cad,parsenet,drpg} require slow per-sample optimization, unsuitable for scalable generation. NURBGen~\cite{nurbgen} takes a promising direction by framing BRep generation as a sequence modeling task using symbolic NURBS, enabling text-to-CAD via LLM fine-tuning. However, generating 3D geometry directly from text without visual grounding leads to low fidelity~\cite{assetgen,marvel} on complex or geometrically precise prompts.

\vspace{0.1cm}
\noindent While industry-standard BRep topology remains 
the gold standard for professional CAD workflows, 
directly training on BReps from multimodal inputs is 
neither scalable nor viable, as the above limitations 
collectively prevent gradient-based learning on 
large-scale BRep data. Inspired by recent progress in 3D generation~\cite{assetgen,clay}, we argue that scalable CAD synthesis requires a decoupled two-stage pipeline: a first stage that learns generalizable 3D shape from large-scale unstructured data, and a second stage that recovers fine-grained CAD topology from these intermediate representations. DreamCAD addresses the first 
stage via $C^0$-continuous Bézier patches with differentiable tessellation, enabling 
point-level supervision on large-scale 3D meshes without CAD-specific annotations.

\vspace{0.1cm}
\noindent \textbf{Multimodal CAD Datasets.}
Unlike 3D vision, where large-scale multimodal datasets such as Objaverse~\cite{objaverse} and MARVEL-40M+~\cite{marvel} have driven progress in text-to-3D generation, the CAD domain remains limited by multimodal data scarcity.
Existing CAD datasets, such as Text2CAD-160K~\cite{text2cad}, contain design histories with text captions but are small in scale. Moreover, large-scale CAD datasets such as ABC-1M~\cite{abc} and Automate-440K~\cite{automate} lack textual or visual descriptions, restricting their use for multimodal generative learning.
Although recent advances in automated 3D captioning, including Cap3D~\cite{cap3d} and MARVEL-40M+~\cite{marvel}, have enabled large-scale annotations for meshes, there are still no comparable resources for BRep models. To bridge this gap, we introduce \textbf{CADCap-1M}, a dataset of over 1M high-quality text descriptions for CAD models automatically generated using GPT-5, for scalable training and evaluation for text-to-CAD research.

\section{Preliminaries}

In this section, we briefly review the CAD representation used in the DreamCAD architecture. Among the various parametric surface formulations, Bézier and NURBS are the most widely adopted in modern CAD modeling. In our implementation, we choose bicubic rational Bézier surfaces due to their conceptual simplicity and analytical tractability. Furthermore, rational Bézier surfaces can be viewed as a special case of NURBS surfaces, making them naturally compatible with standard CAD operations.

\vspace{0.1cm}
\noindent \textbf{Rational Bézier Surface.}
A rational Bézier surface of degree $(n,m)$ is defined by $(n+1)\times (m+1)$ control points $C=\{c_{ij}\}$ and non-negative weights $W=\{w_{ij}\}$. The surface $S(u,v)$ is evaluated in $uv$ domain as:
\begin{equation}
S(u,v) =
\frac{\sum_{i,j} B_i^n(u) B_j^m(v) w_{ij} c_{ij}}
{\sum_{i,j} B_i^n(u) B_j^m(v) w_{ij}},
\end{equation}
where $B_i^n(u)=\binom{n}{i}u^i(1-u)^{n-i}$ and $B_j^m(v)=\binom{m}{j}v^j(1-v)^{m-j}$ are Bernstein basis functions and $(u,v) \in [0,1]^2$. For a bicubic case $n=m=3$. 
This formulation is differentiable with respect to both control points and weight. It is worth noting that the weights $w_{ij}$ must remain non-negative, as negative weights can lead to degenerate or invalid surface evaluations.

\begin{figure}[t]
    \centering
    \includegraphics[width=0.8\linewidth]{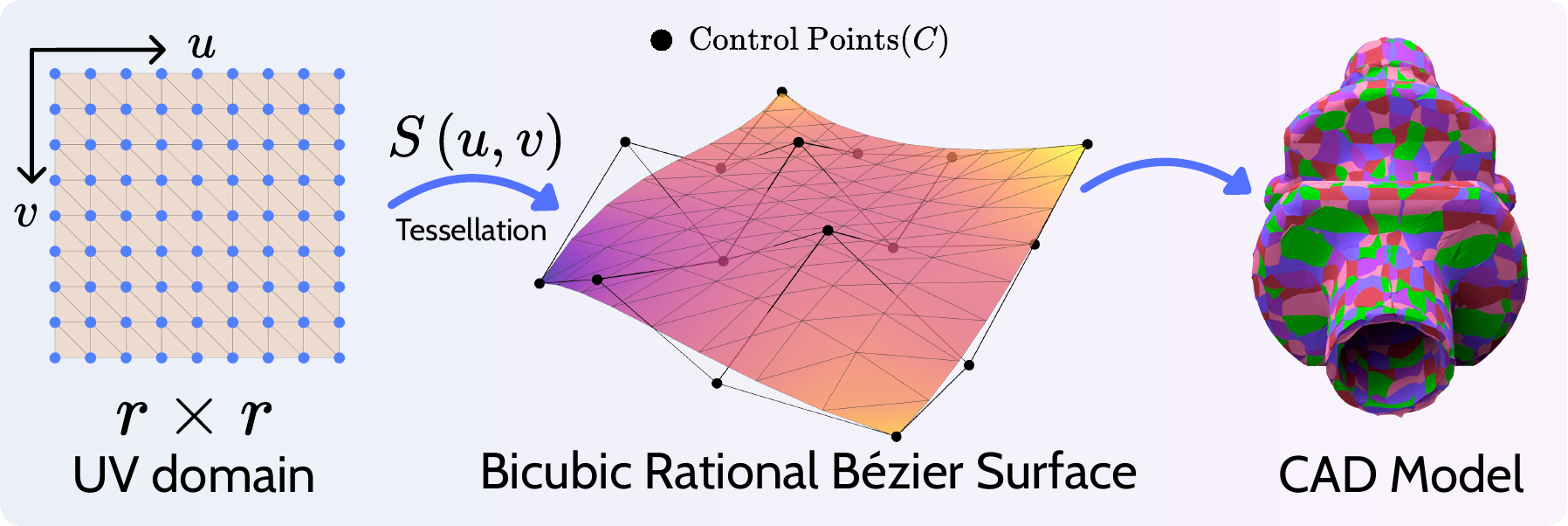}
    \caption{Bézier surface representation and differentiable tessellation.}
    \label{fig:tessellation}
\end{figure}

\begin{figure*}[t]
    \centering
    \includegraphics[width=1\linewidth]{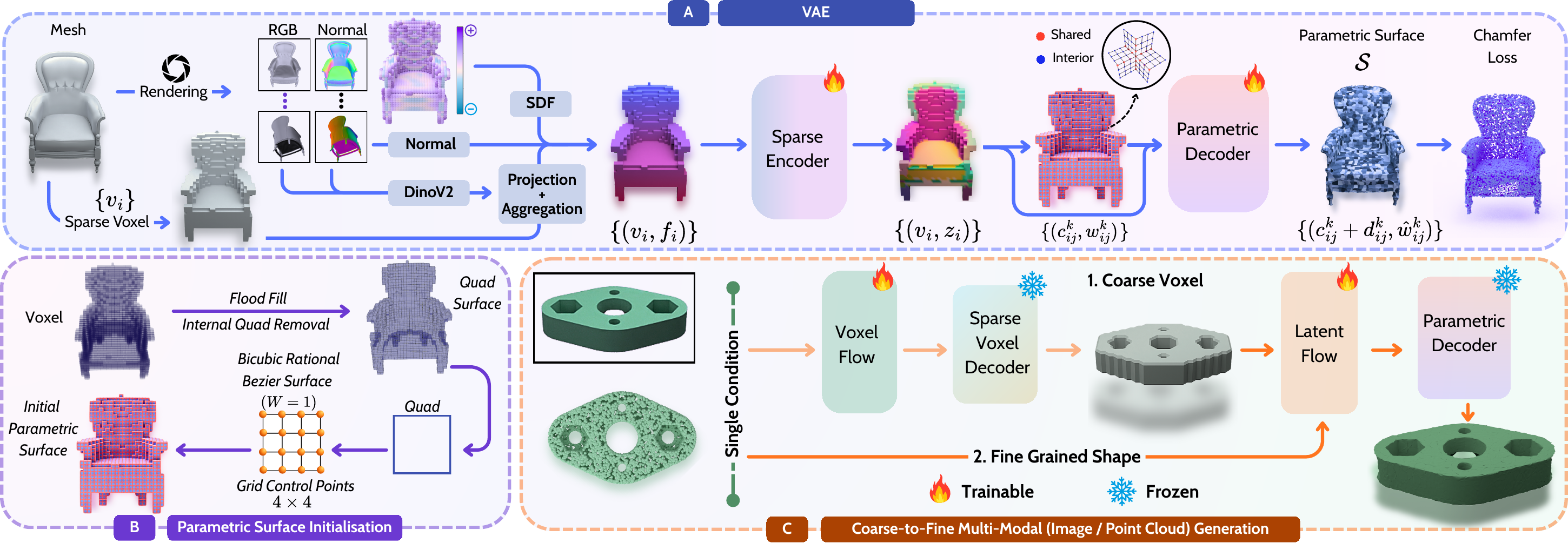}
    \caption{\textbf{DreamCAD Overview:} \textcolor{vae}{\textbf{(A). Sparse Transformer VAE}} from an input mesh, active voxels $v_i$ are generated with local features $f_i$, from DINOv2~\cite{dinov2} embeddings, normal images, and SDF values and encodes it to generate structured latents $z_i$. These are then decoded into parametric (rational bézier) surfaces and optimized using Chamfer loss.  \textcolor{initial_surface}{\textbf{(B). Initial $C^0$-continuous Parametric Surface}} generation from sparse voxels via flood-fill and quad conversion using grid control points and unit weights. \textcolor{generation}{\textbf{(C). Multi-modal CAD generation}} from images, or points using a flow-matching framework from coarse voxel grid to parametric surface refinement.}
    \label{fig:architecture}
\end{figure*}

\vspace{0.1cm}
\noindent \textbf{Differentiable Mesh Generation.}
Given a set of Bézier patches $\{S_k\}_{k=1}^K$, we generate meshes through differentiable tessellation~\cite{drpg} (Figure~\ref{fig:tessellation}). Each patch $S_k(u,v)$ is evaluated on the $uv$ domain by uniformly sampling $(u,v)$ points on a grid of resolution $r\times r$. Adjacent points in this grid define quadrilateral cells, which are then split into triangles to form a locally consistent mesh. Neighboring Bézier patches are merged along shared boundaries to ensure $C^0$ connectivity. Since $S(u,v)$ is differentiable with respect to its $C$ and $W$, the entire tessellation process supports end-to-end gradient-based optimization.

 \vspace*{-.3\baselineskip}
\section{Methodology}
 \vspace*{-.3\baselineskip}
As shown in Figure~\ref{fig:architecture}, \textbf{DreamCAD} adopts a multi-stage generative pipeline. Section~\ref{subsec:vae} presents the VAE module, which encodes 3D shapes into compact latent representations and decodes them into a parametric surface. Section~\ref{subsec:cadcap} details the automatic captioning pipeline. Finally, Section~\ref{subsec:conditional} introduces the conditional generation framework, which uses a coarse-to-fine strategy: sparse voxels are first generated from the input condition, followed by reconstruction of fine-grained parametric surfaces.

\subsection{Latent Encoding} \label{subsec:vae}

\noindent \textbf{Sparse Voxel Representation.} An effective 3D encoder requires a compact yet structured representation. While point clouds are widely used~\cite{hunyuan3d,step1x3d}, they lack spatial regularity for continuous CAD generation. We instead adopt a sparse voxel representation enriched with local visual features~\cite{trellis}. We first voxelize each input mesh to $32^3$ resolution, generating active voxels $\{v_i\}_{i=1}^N$. 
To preserve fine geometric details, each active voxel is augmented with visual cues. We render 150 RGB and normal views of the mesh from different camera angles, extract DINOv2~\cite{dinov2} embeddings, and project each voxel center $v_i$ to obtain per-view RGB and normal features. Averaging across views gives:
\begin{equation}
\begin{aligned}
p(v_i) = \frac{1}{150} \sum_{j=1}^{150} 
\Big[\, 
    \texttt{proj}(v_i, E^r_j);\;
     \texttt{proj}(v_i, E^n_j)
\,\Big],
\end{aligned}
\end{equation}
\noindent where $\texttt{proj}$ is the projection operator and $p(v_i)$ is the mean feature vector, $E^r_j$ and $E^n_j$ are DINO embedding maps from the $j$-th RGB and normal views. 
We further include per-view normals $n(v_i)\in\mathbb{R}^{150\times3}$, voxel centers $c(v_i)\in\mathbb{R}^3$, and signed distance values $s(v_i)\in\mathbb{R}$ to encode geometry and surface proximity. The final voxel feature is:
\vspace*{-.4\baselineskip}
\begin{equation}
f_i = [p(v_i); n(v_i); c(v_i); s(v_i)], \quad f_i\in\mathbb{R}^{2502}
\end{equation}
These features are processed by a sparse Transformer encoder~\cite{trellis} to produce structured latents $\{(v_i, z_i)\}_{i=1}^{N}$.

\vspace{0.1cm} 
\noindent \textbf{Initial Parametric Quad Generation.} VAE Decoder's goal is to reconstruct the 3D shape as a set of Bézier patches from the structured latent codes. Each patch requires $16$ control points and corresponding weights. Directly generating these parameters from latents $z_i$ leads to disconnected or overlapping patches, as losses alone cannot enforce $C^0$ continuity between two adjacent patches. 

\vspace{0.1cm}
\noindent To mitigate this, we generate an initial parametric quad surface from sparse voxels (Fig.~\ref{fig:architecture}.B). First, we generate a surface mesh from the sparse voxels using a flood-fill algorithm, which removes the internal quads from the voxels. We then convert each quad into a bicubic rational bézier patch by uniformly sampling a $4 \times 4$ grid points using bilinear interpolation of its four corners.

\vspace{0.1cm}
\noindent All control points start with unit weights, and adjacent patches share boundary control points (``\tikz[baseline=-0.7ex]\draw[red,fill=red] (0,0) circle (2.5pt);~Shared''
 in Figure~\ref{fig:architecture}.A) to ensure $C^0$ continuity. The resulting surface $S$ is represented as:
 \begin{equation}
\begin{aligned}
\mathcal{S} 
    &= \{s_k = (c^k_{ij};\, w^k_{ij})\} \in \mathbb{R}^{N_f\times 16\times4}, \quad k \in \mathbb{Z}_{[0, N_f]},\;
    i,j \in \{0,1,2,3\},
\end{aligned}
\end{equation}
where $N_f$ denotes the number of patches, $s_k$ denotes the $k^{th}$ patch and $i,j$ are the indices for both control points and weights. In practice, we observe that the number of patches per shape remains low, typically $N_f \ll 10000$.

\vspace{0.1cm} 
\noindent \textbf{Parametric Surface Decoder.} The decoder refines the initial surface $\mathcal{S}$ by using structured latent features to predict local adjustments for each patch. For the $(i,j)$-th control point of the $k$-th patch, it predicts a deformation $d^k_{ij}$ and a weight update $\hat{w}^k_{ij}$. Unconstrained $d^k_{ij}$ often causes degenerate geometries such as spikes or self-intersections, which are hard to recover during optimization. Therefore, to stabilize training, we bound the deformation within a local neighborhood using $c^k_{ij} \leftarrow c^k_{ij} + \tanh(d^k_{ij})$. We ensure positive weights via the softplus function $\log(1 + e^{\hat{w}^k_{ij}})$. For shared boundary control points, we enforce $C^0$ 
continuity by uniformly averaging the predicted deformations and weight updates from all patches sharing that point, ensuring all adjacent patches converge to the same boundary position.

\begin{figure}[t]
    \centering
    \includegraphics[width=1\linewidth]{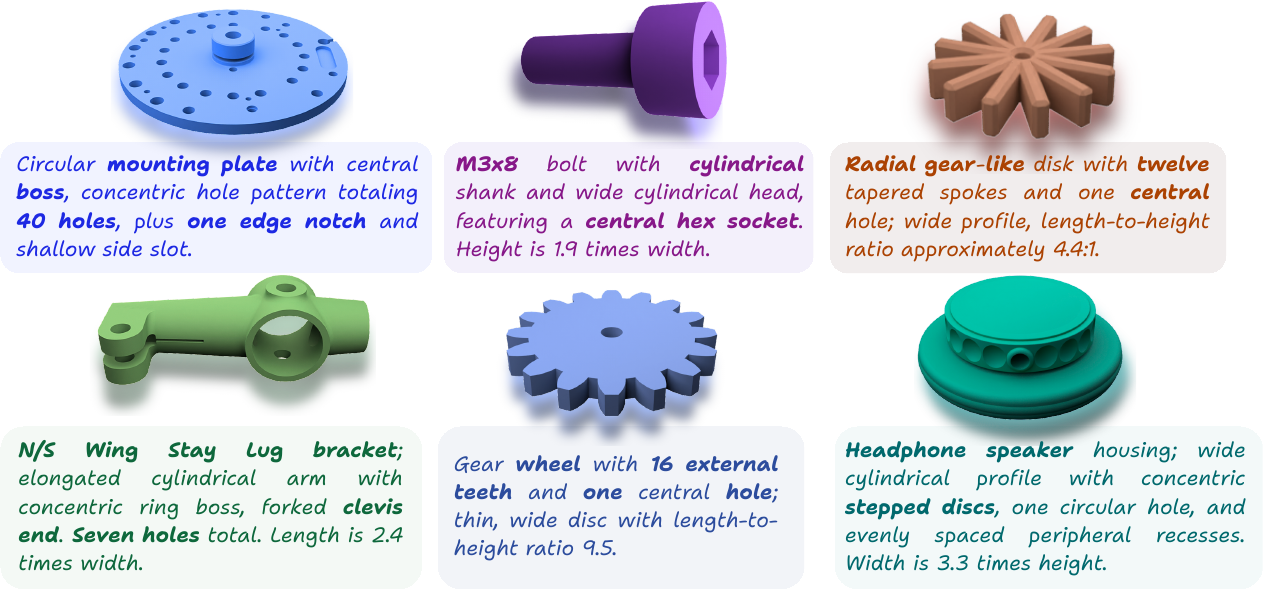}
    \caption{Examples of metadata-augmented captions from CADCap-1M showing object type, part names, and hole counts.}
    \vspace*{-.4\baselineskip}
    \label{fig:caption}
    \vspace*{-.4\baselineskip}
\end{figure} 

%
\vspace{0.1cm}
\noindent We then tessellate the deformed surface into a mesh and compute Chamfer distance (CD) loss between sampled surface points ($\mathcal{X}_{d}$) and the target point cloud ($\mathcal{X}_{g}$).
\noindent The overall training objective is:
\vspace*{-.4\baselineskip}
\begin{equation}\label{eq:vae_loss}
\begin{aligned}
\mathcal{L}= & \lambda_{cd}\texttt{CD}(\mathcal{X}_{g},\mathcal{X}_{d}) + \lambda_{g1}\texttt{G1}(\mathcal{S}_d) +  \lambda_{lp}\texttt{Laplacian}(\mathcal{M}_d) +  \lambda_{kl}D_{KL},
\end{aligned}
\end{equation}

where $\mathcal{M}_d$ is the tessellated mesh from $\mathcal{S}_d$, \texttt{G1}~\cite{g1} enforces tangent continuity between two adjacent patches, \texttt{Laplacian}~\cite{laplacian} ensures smoothness and $D_{\text{KL}}$ regularizes the VAE latent space using KL divergence loss. 


\vspace*{-.4\baselineskip}
\subsection{CADCap-1M Dataset} \label{subsec:cadcap}
Recent progress in text-to-3D generation has been driven by large-scale captioned datasets such as MARVEL-40M+~\cite{marvel} and Cap3D~\cite{cap3d}, yet no comparable resource exists for \textit{text-to-CAD}. To progress the field, we introduce CADCap-1M, comprising $1$M+ high-quality captions for CAD models from ABC~\cite{abc}, Automate~\cite{automate}, CADParser~\cite{cadparser}, Fusion360~\cite{fusion360}, ModelNet~\cite{modelnet}, 3D-Future~\cite{3d-future}. For each model, we render four orthographic views using Blender~\cite{blender} and prompt GPT-5~\cite{gpt5} to generate concise descriptions. Prompts are augmented with metadata from the original CAD files, such as model names which are optionally extracted from \textit{.step} files, number of holes, which is computed using~\cite{milnor1997topology}, and relative dimensions (\textit{length-to-width},\textit{ width-to-height}, or \textit{length-to-height} ratios). This metadata-augmented prompting substantially reduces hallucinations and improves geometric accuracy and linguistic quality~\cite{marvel,text2cad}. As a result, CADCap-1M produces more shape-centric and structure-aware captions (e.g., “\textit{M3x8 bolt ..}”, “\textit{.. mounting plate .. 40 circular holes ..}”), as illustrated in Figure~\ref{fig:caption}. 




\subsection{Conditional CAD Generation} \label{subsec:conditional}
We now describe the process of conditional CAD generation from text, images, and point clouds. 


\vspace{0.1cm}

\noindent \textbf{Generation Pipeline.} As shown in Figure~\ref{fig:architecture}.C, DreamCAD adopts a coarse-to-fine generation pipeline with two Flow Transformer Decoders optimized via flow matching (FM) objectives~\cite{flow-matching}. FM learns to estimate velocity field $v_\theta(x_t, t)$ for transforming samples from a prior distribution $x_0\sim p_0$ to the target distribution $x_1\sim p_1$ through the loss $\mathcal{L}_{\text{FM}} = \mathbb{E}_{x_0, x_1, t} [| v_\theta(x_t, t) - (x_1 - x_0) |_2^2]$.

\vspace{0.1cm}
\noindent In the first stage, we generate a coarse voxel grid from the input condition by producing a low-resolution latent structure through a lightweight VAE trained to reconstruct voxel grids, following~\cite{trellis}. The pretrained VAE decoder efficiently upsamples this latent grid into a full-resolution voxel representation. In the second stage, we generate local SLAT features ($z_i$) for each active voxel ($v_i$) using the predicted voxel grid and conditioning input. Finally, the pretrained parametric surface decoder transforms $\{(v_i,z_i)_{i=1}^{N}\}$ into the final parametric surface $\mathcal{S}$. For conditional embedding, we use modality-specific encoders: DINOv2~\cite{dinov2} for images and PointNet++~\cite{pointnet++} for point clouds. We use pretrained weights for DINOv2 while training PointNet++ jointly with the Flow Transformer models.

\vspace{0.1cm}
\noindent \textbf{Text-to-CAD Generation.} While our framework supports text conditioning, training a direct text-to-3D model typically leads to slow convergence~\cite{assetgen} and low prompt fidelity~\cite{marvel}, due to the lack of explicit spatial and geometric cues in textual input. Following standard text-to-3D practices~\cite{marvel,clay,assetgen,instant3d}, we adopt a two-stage approach: \textit{text-to-image} followed by \textit{image-to-CAD}. We fine-tune Stable Diffusion 3.5-2B~\cite{stable-diff3} on the CADCap-1M dataset to align its output image distribution with that of the \textit{image-to-CAD} model~\cite{marvel}. The resulting images are then used to condition our pretrained \textit{image-to-CAD} model. This significantly enhances prompt fidelity with reduced training cost. Notably, pretrained \textit{text-to-image} models~\cite{flux,stable-diff3} often fail to preserve numerically constrained features such as hole or gear-tooth counts, underscoring the importance of fine-tuning for accurate \textit{text-to-image} generation for the CAD domain.

\vspace*{-.3\baselineskip}
\section{Experiments} 
\vspace*{-.3\baselineskip}

We evaluate DreamCAD on conditional CAD generation from text, image, and point inputs in Section~\ref{sec:generation}. Section~\ref{sec:caption_quality} analyzes the caption quality of the CADCap-1M dataset, and Section~\ref{sec:ablation} presents ablation studies on key architectural and training design choices.

\vspace{0.1cm}
\noindent \textbf{Datasets.} We curate over 1M 
high-quality CAD models from 10 public datasets 
(Table~\ref{tab:dreamcad_training_dataset}) to train 
the DreamCAD VAE, converting all BRep-only datasets 
to meshes using OpenCascade. Although large-scale datasets like 
ABC and Automate contain BReps, training directly on 
BReps is impractical at scale as BRep topology is 
discrete and non-differentiable. We filter low-quality and trivial 
primitives from ABC~\cite{abc} and 
Automate~\cite{automate}, splitting into 95\% training, 
2.5\% validation, and 2.5\% testing. For 
\textit{text-to-CAD}, we use CADCap-1M and 
MARVEL~\cite{marvel} Level-5 captions for 
ShapeNet~\cite{shapenet}, Toys4K~\cite{toys4k}, and 
ABO~\cite{3d-future}. For \textit{image-to-CAD}, we 
render four orthographic views (front, back, left, 
right) in Blender~\cite{blender} and randomly sample 
one view per epoch. For \textit{point-to-CAD}, we 
normalize point clouds to $[-0.5, 0.5]$ and augment 
with surface normals in 50\% of batches. We discuss more details in the supplementary.

\begin{figure}[t]
    \centering
    \includegraphics[width=1\linewidth]{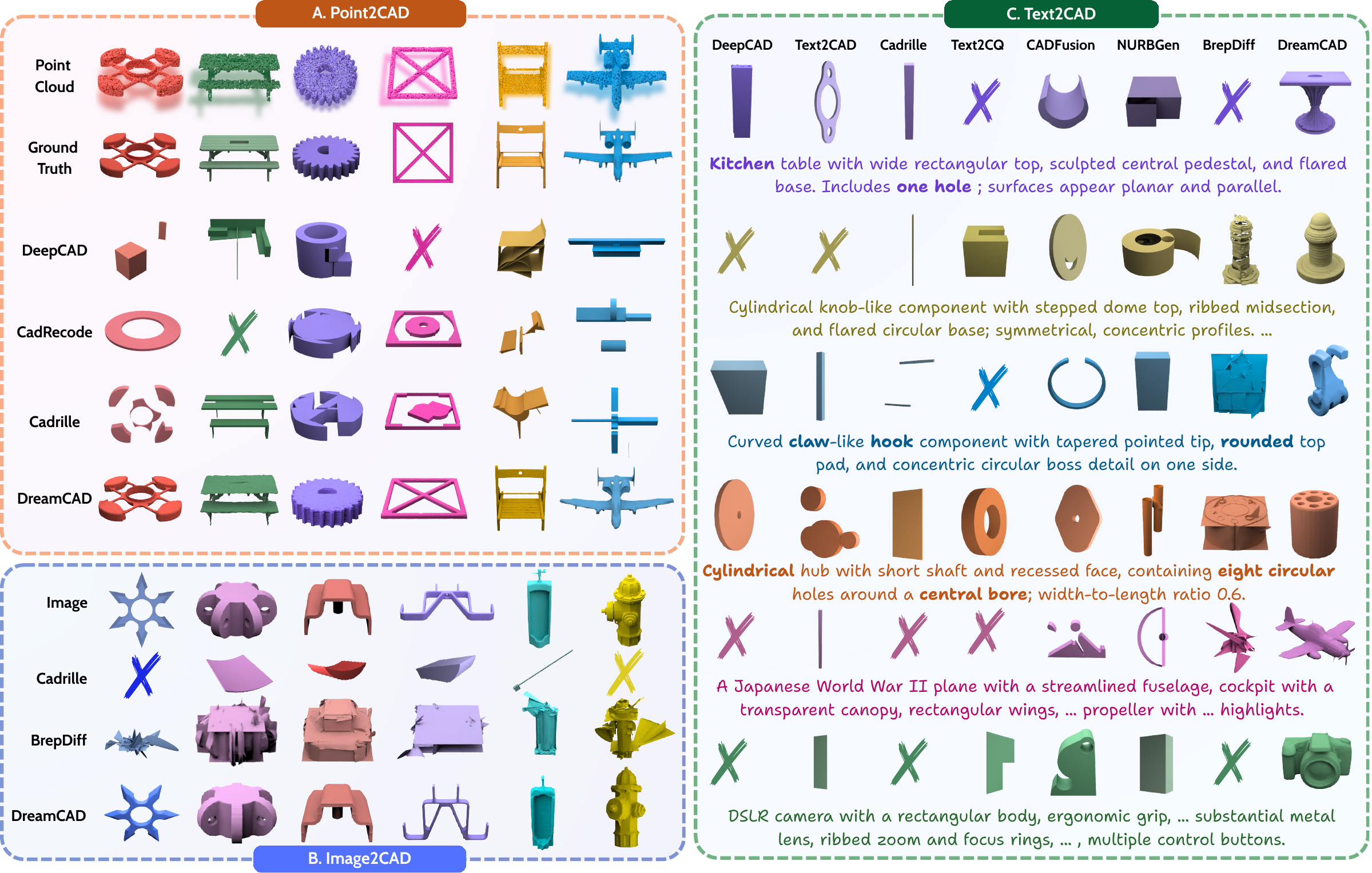}
    \caption{Qualitative comparison on \textcolor{point2cad}{\textbf{Point2CAD}} (Top-Right), \textcolor{image2cad}{\textbf{Image2CAD}} (Bottom-Left) and \textcolor{text2cad}{\textbf{Text2CAD}} (Right) tasks. For each task, the first four examples are from the ABC dataset, while the last two from Objaverse dataset. \xmark $\,$ indicates invalid models.}
    \label{fig:qualitative}
    \vspace*{-.5\baselineskip}
    \label{fig:generation}
    \vspace*{-.5\baselineskip}
\end{figure}

\begin{table*}[t]
\def\arraystretch{1.2}%
\caption{\textbf{Quantitative comparison} across the Point2CAD, Img2CAD, and Text2CAD tasks on the ABC and Objaverse datasets. For readability, F1 is scaled by $10^2$ and CD, JSD, and MMD by $10^3$. For text- and image-to-CAD, GPT and User ratings measure visual alignment.}
\vspace*{-.7\baselineskip}
\resizebox{1\textwidth}{!}{
\begin{tabular}{ll:cccccc:cccccc}
\hline
\multicolumn{1}{c}{\multirow{2}{*}{Task}} & \multicolumn{1}{c}{\multirow{2}{*}{Models}} 
& \multicolumn{6}{c}{ABC} & \multicolumn{6}{c}{Objaverse} \\ 
\multicolumn{1}{c}{} & \multicolumn{1}{c}{} 
& \cellcolor{ubc!20}F1 $\uparrow$ & \cellcolor{ubc!20}NC $\uparrow$ & \cellcolor{blue!10}CD $\downarrow$ & \cellcolor{blue!10}HD $\downarrow$ & \cellcolor{blue!10}JSD $\downarrow$ & \cellcolor{blue!10}MMD $\downarrow$ 
& \cellcolor{ubc!20}F1 $\uparrow$ & \cellcolor{ubc!20}NC $\uparrow$ & \cellcolor{blue!10}CD $\downarrow$ & \cellcolor{blue!10}HD $\downarrow$ & \cellcolor{blue!10}JSD $\downarrow$ & \cellcolor{blue!10}MMD $\downarrow$ \\ \hline
\addlinespace[2pt]

\multirow{4}{*}{\rotatebox{90}{\small \textbf{Point2CAD}}} 
& DeepCAD~\cite{deepcad} & 19.31 & 0.49 & 51.10 & 0.37 & 783.94 & 29.63 & 7.05 & 0.48 & 320.33 & 0.41 & 855.14 & 34.62 \\
& CAD-Recode~\cite{cadrecode} & 75.99 & 0.79 & 3.73 & 0.13 & 271.89 & 2.94 & 53.24 & 0.66 & 7.92 & 0.19 & 479.50 & 6.27 \\
& Cadrille~\cite{cadrille} & 78.86 & 0.80 & 2.98 & 0.12 & 236.10 & 2.51 & 57.49 & 0.67 & 6.28 & 0.17 & 445.23 & 5.24 \\
& \textbf{DreamCAD (Ours)} & \textbf{92.12} & \textbf{0.94} & \textbf{0.93} & \textbf{0.06} & \textbf{96.13} & \textbf{0.84} & \textbf{87.31} & \textbf{0.89} & \textbf{1.25} & \textbf{0.11} & \textbf{189.12} & \textbf{1.86} \\ \hline

\addlinespace[1pt]
\multirow{4}{*}{\rotatebox{90}{\small \textbf{Img2CAD}}} 
& & \multicolumn{1}{c}{\cellcolor{ubc!20}GPT $\uparrow$} & \multicolumn{1}{c}{\cellcolor{ubc!20}User $\uparrow$} & \multicolumn{1}{c}{\cellcolor{blue!10}CD $\downarrow$} & \multicolumn{1}{c}{\cellcolor{blue!10}HD $\downarrow$} & \multicolumn{1}{c}{\cellcolor{blue!10}JSD $\downarrow$} & \multicolumn{1}{c:}{\cellcolor{blue!10}MMD $\downarrow$} & \multicolumn{1}{c}{\cellcolor{ubc!20}GPT $\uparrow$} & \multicolumn{1}{c}{\cellcolor{ubc!20}User $\uparrow$} & \multicolumn{1}{c}{\cellcolor{blue!10}CD $\downarrow$} & \multicolumn{1}{c}{\cellcolor{blue!10}HD $\downarrow$} & \multicolumn{1}{c}{\cellcolor{blue!10}JSD $\downarrow$} & \multicolumn{1}{c}{\cellcolor{blue!10}MMD $\downarrow$} \\ 
& Cadrille~\cite{cadrille} & 7.75 & 5.34 & 111.50 & 0.49 & 909.92 & 68.97 & 1.10 & 0.45 & 99.91 & 0.48 & 913.24 & 58.02 \\
& BRepDiff~\cite{brepdiff} & 16.13 & 17.63 & 20.69 & 0.28 & 662.97 & 13.89 & 18.12 & 16.63 & 57.51 & 0.43 & 875.68 & 27.83 \\
& \textbf{DreamCAD (Ours)} & \textbf{76.12} & \textbf{77.03} &\textbf{ 4.12} & \textbf{0.17} & \textbf{412.31} & \textbf{6.31} & \textbf{80.78} & \textbf{82.92} & \textbf{20.16} & \textbf{0.27} & \textbf{541.81} & \textbf{13.41} \\ \hline
\addlinespace[1pt]
\multirow{10}{*}{\rotatebox{90}{\small \textbf{Text2CAD}}} 
& & \multicolumn{1}{c}{\cellcolor{ubc!20}GPT $\uparrow$} & \multicolumn{1}{c}{\cellcolor{ubc!20}User $\uparrow$} & \multicolumn{1}{c}{\cellcolor{blue!10}CD $\downarrow$} & \multicolumn{1}{c}{\cellcolor{blue!10}HD $\downarrow$} & \multicolumn{1}{c}{\cellcolor{blue!10}JSD $\downarrow$} & \multicolumn{1}{c:}{\cellcolor{blue!10}MMD $\downarrow$} & \multicolumn{1}{c}{\cellcolor{ubc!20}GPT $\uparrow$} & \multicolumn{1}{c}{\cellcolor{ubc!20}User $\uparrow$} & \multicolumn{1}{c}{\cellcolor{blue!10}CD $\downarrow$} & \multicolumn{1}{c}{\cellcolor{blue!10}HD $\downarrow$} & \multicolumn{1}{c}{\cellcolor{blue!10}JSD $\downarrow$} & \multicolumn{1}{c}{\cellcolor{blue!10}MMD $\downarrow$} \\ 
& DeepCAD~\cite{deepcad} & 0.49 & 0.40 & 86.54 & 0.44 & 887.69 & 37.66 & 0.00 & 0.00 & 80.68 & 0.45 & 903.40 & 37.34 \\
& Text2CAD~\cite{text2cad} & 1.11 & 2.46 & 82.22 & 0.41 & 852.68 & 40.65 & 0.22 & 0.36 & 93.96 & 0.47 & 896.22 & 44.79 \\
& Cadrille~\cite{cadrille} & 0.91 & 0.34 & 155.80 & 0.53 & 957.51 & 96.89 & 0.04 & 0.00 & 162.16 & 0.55 & 961.53 & 76.39 \\
& Text2CQ~\cite{text2cadquery} (Qwen3B) & 0.01 & 0.00 & 68.15 & 0.39 & 829.72 & 37.50 & 0.00 & 0.00 & 83.43 & 0.44 & 890.89 & 47.62 \\
& Text2CQ (GPT2L) & 0.02 & 0.00 & 71.27 & 0.39 & 838.54 & 40.35 & 0.00 & 0.00 & 84.85 & 0.47 & 891.96 & 53.70 \\
& Text2CQ (CodeGPT) & 0.94 & 1.00 & 77.91 & 0.41 & 850.23 & 43.71 & 0.24 & 0.12 & 86.75 & 0.46 & 879.57 & 58.48 \\
& CADFusion~\cite{cadfusion} & 2.35 & 2.76 & 56.36 & 0.31 & 789.12 & 27.54 & 2.67 & 2.35 & 81.03 & 0.43 & 853.43 & 58.11 \\
& NURBGen~\cite{nurbgen} & 4.21 & 4.44 & 50.84 & 0.32 & 800.46 & 29.53 & 8.18 & 7.28 & 73.54 & 0.41 & 839.08 & 38.69 \\
& BRepDiff~\cite{brepdiff} & 4.34 & 3.20 & 54.12 & 0.38 & 812.31 & 34.72 & 6.33 & 6.41 & 74.32 & 0.38 & 808.19 & 41.13 \\
& \textbf{DreamCAD (Ours)} & \textbf{85.62} & \textbf{85.40} & \textbf{20.32} & \textbf{0.14} & \textbf{734.92} & \textbf{19.43} & \textbf{82.32} & \textbf{83.48} & \textbf{34.61} & \textbf{0.28} & \textbf{698.21} & \textbf{28.14} \\ \hline
                                                
\end{tabular}
}
\vspace*{-.6\baselineskip}
\label{tab:quantitative}
\vspace*{-.6\baselineskip}
\end{table*}

\vspace{0.1cm}
\noindent \textbf{Implementation.} We train DreamCAD VAE for $700$k steps for $3$ weeks with batch size $32$ using mixed-precision, AdamW~\cite{adamw} with learning rate $5\times10^{-5}$ and weight decay $1\times10^{-4}$. Both VAE encoder and decoder contain 8 Transformer layers with voxel latent dimension $z_i=8$. We empirically set the loss weights in Eq.~\ref{eq:vae_loss} to $\lambda_{cd}=10^2$, $\lambda_{g1}=5{\times}10^{-3}$, and $\lambda_{lp}=1$. During training, points for CD loss increase from 16K to 100K via sigmoid scheduling and tessellation resolution $r$ increases from $(4,4)$ to $(16,16)$. We train both coarse and fine-grained flow Transformer decoders for $500$k steps with $10\%$ condition dropout. Pretrained image embeddings have a dimensionality of 1536. During inference, we use 50 steps and set the classifier free guidance scale to 7.5. For Stable Diffusion fine-tuning (text to image), we apply LoRA~\cite{lora} with rank and $\alpha{=}4$, trained for $300$k steps. Inference takes $\sim$15s for \textit{image}- and \textit{point-to-CAD}, and $\sim$30s for \textit{text-to-CAD}.  

\vspace*{-.5\baselineskip}
\subsection{Multimodal Generation Evaluation}\label{sec:generation}

\noindent \textbf{Experimental Setup.}
We evaluate DreamCAD on two datasets: ABC~\cite{abc} and Objaverse~\cite{objaverse}, each containing 15K samples. ABC serves as the in-distribution benchmark, while Objaverse is used for out-of-distribution (OOD) evaluation to measure generalization. Because Objaverse includes many free-form and organic objects uncharacteristic of CAD geometry, we filter its test set using MARVEL~\cite{marvel} captions containing CAD-specific keywords.

\vspace{0.1cm}
\noindent \textbf{Baselines.}
As the first CAD generative framework trained on large-scale unstructured 3D data, DreamCAD has no direct counterparts. We therefore compare against representative design-history and UV-based methods across all tasks. For \textit{text-to-CAD}, we include Text2CAD~\cite{text2cad}, Text2CQ~\cite{text2cadquery}, Cadrille~\cite{cadrille}, DeepCAD~\cite{deepcad}, CADFusion~\cite{cadfusion}, NURBGen~\cite{nurbgen} and BRepDiff~\cite{brepdiff}. For CADFusion~\cite{cadfusion}, we generate 5 outputs per sample as per the official implementation. We observe this drastically reduces the invalidity ratio. For the \textit{image-to-CAD} task, we compare DreamCAD against BRepDiff and Cadrille while for the \textit{point-to-CAD} we additionally include DeepCAD and CAD-Recode~\cite{cadrecode}. DeepCAD is trained for 100 epochs on both \textit{point} and \textit{text}-to-CAD tasks on DeepCAD and Text2CAD datasets, while  BRepDiff is trained for \textit{image-to-cad} for 6k epochs on ABC. Since the training schemes introduce a data-scale disparity, we additionally report results on the DeepCAD dataset in the supplementary.

\vspace{0.1cm}
\noindent \textbf{Metrics.} 
We evaluate all tasks using both geometric and perceptual metrics. Geometric fidelity is assessed by Chamfer Distance (CD), Hausdorff Distance (HD), Jensen–Shannon Divergence (JSD), Minimum Matching Distance (MMD), Normal Consistency (NC), and F1 score. All geometric metrics are computed on $8192$ uniformly sampled points normalized within a unit cube centered at the origin. For \textit{text}- and \textit{image-to-CAD}, visual alignment is measured on $5\text{k}$ and $1\text{k}$ samples through GPT-5~\cite{gpteval3d} and user studies by 14 CAD-experts, respectively. In both settings, GPT-5 and human evaluators are shown multi-view renderings of reconstructions from all baselines and DreamCAD, and asked to select the model best matching the input (\textit{text} or \textit{image}). The same expert group is used consistently across all user studies.



\vspace{0.1cm}
\noindent \textbf{Results.} As shown in Table~\ref{tab:quantitative}, DreamCAD achieves state-of-the-art results across all three modalities in \textit{point-to-CAD}, \textit{image-to-CAD}, and \textit{text-to-CAD}. In the easier \textit{point-to-CAD} task, it outperforms baselines on ABC by a large margin, reducing CD by up to 68\% and 75\% over Cadrille and CAD-Recode and improving F1 scores by 17\% and 21\%, respectively. Similar gains are observed across other metrics and the Objaverse dataset as well. As illustrated in Figure~\ref{fig:qualitative}.A, DreamCAD accurately reconstructs complex geometries such as gear wheels (Ex.~3) and chair (Ex.~5), plane (Ex.~6), whereas prior methods capture only the coarse shapes. 

\vspace{1mm}
\noindent For \textit{image-to-CAD}, DreamCAD achieves over 75\% preference in both GPT and human evaluations on ABC and Objaverse, improving CD by 80\% and 58\% and MMD by 54\% and 52\% over BRepDiff, respectively. As shown in Figure~\ref{fig:qualitative}.B, DreamCAD produces high-fidelity reconstructions from single images via its coarse-to-fine pipeline.


\vspace{1mm}
\noindent For the most challenging \textit{text-to-CAD} task, DreamCAD attains over 80\% preference in both GPT and user studies, substantially outperforming all baselines. The closest competitor, BRepDiff, which takes as input the same images generated by our finetuned SD~3.5, remains far behind in reconstruction quality. Geometrically, DreamCAD achieves a 62\% reduction in CD compared to the second-best method NURBGen in the ABC dataset. As shown in Figure~\ref{fig:qualitative}.C, it exhibits strong prompt fidelity, accurately reconstructing intricate shapes (Ex.~1: table, Ex.~6: camera) and numerically constrained features (Ex.~4: hole counts). Design-history-based models fail beyond simple primitives, while NURBGen's lower prompt fidelity demonstrates that end-to-end text-to-CAD without visual grounding struggles to capture precise geometric details. Furthermore, BRepDiff's post-processing stage occasionally produces disconnected BRep faces and visible spike artifacts from misplaced UV grid points, a limitation of 
grid-based representations that prevents reliable $C^0$ continuity.


\begin{table}[t]
\def\arraystretch{1.2}%
\centering
\caption{Ablation studies on regularization (left), and voxel resolution choice (right).}
\vspace*{-.9\baselineskip}
\vspace*{-.4\baselineskip}
\label{tab:ablation_combined}
\begin{subtable}[t]{0.5\textwidth}
\centering
\caption{Impact of regularizers on VAE performance.}
\label{tab:regularization}
\resizebox{\columnwidth}{!}{
\begin{tabular}{lccc}
\hline
\cellcolor{blue!10}Model    & \multicolumn{1}{c}{\cellcolor{blue!10}CD ($\times 10^3$) $\downarrow$} & \multicolumn{1}{c}{\cellcolor{blue!10}Lap ($\times 10^3$)$\downarrow$} & \multicolumn{1}{c}{\cellcolor{blue!10}HD$\downarrow$}  \\ \hline
No Regularization    &                        \textbf{0.0210} &                         0.0073&                        \textbf{0.02}\\
 + G1     &                         0.0230&                         0.0064&                        0.022\\
+ Lap    &                        0.0225&                         0.0022&                        0.022\\
 + G1 + Lap  &                        0.0259&                         \textbf{0.0020}&                        0.024\\ \hline
\end{tabular}
}

\end{subtable}
\hfill
\begin{subtable}[t]{0.45\textwidth}
\centering
\caption{Ablation on voxel-grid resolution}
\resizebox{\columnwidth}{!}{
\begin{tabular}{lcccc}
\hline
Voxel-Grid          & 24      & 32      & 48      & 64       \\ \hline
\#patches           & 1434.32 & 2546.18 & 7720.48 & 10179.34 \\
CD $(\times 10^{3})$ & 0.0231  & 0.011   & 0.0109  & 0.0105   \\ 
\hline
\end{tabular}
}
\label{tab:ablation}
\end{subtable}
\vspace*{-.5\baselineskip}
\end{table}

\vspace*{-.3\baselineskip}
\subsection{Caption Quality}\label{sec:caption_quality}
\vspace*{-.3\baselineskip}
As \textbf{CADCap-1M} is the first large-scale captioning dataset for CAD models, no existing benchmark enables direct comparison. We therefore evaluate caption quality through both user studies and GPT-5 assessment on $1k$ and $5k$ samples respectively. Given four rendered views, metadata, and the caption, evaluators rate both geometric and semantic accuracy. Overall, $95.8$\% (user) and  $98.31$\% (GPT-5) of captions are judged correct, including precise identification of part names and hole counts. This validates the reliability of our metadata-augmented prompting. We provide more statistical analyses in the supplementary material. 


\vspace*{-.3\baselineskip}
\subsection{Ablation Study}\label{sec:ablation}

\begin{figure}[t]
    \centering
\includegraphics[width=0.65\linewidth]{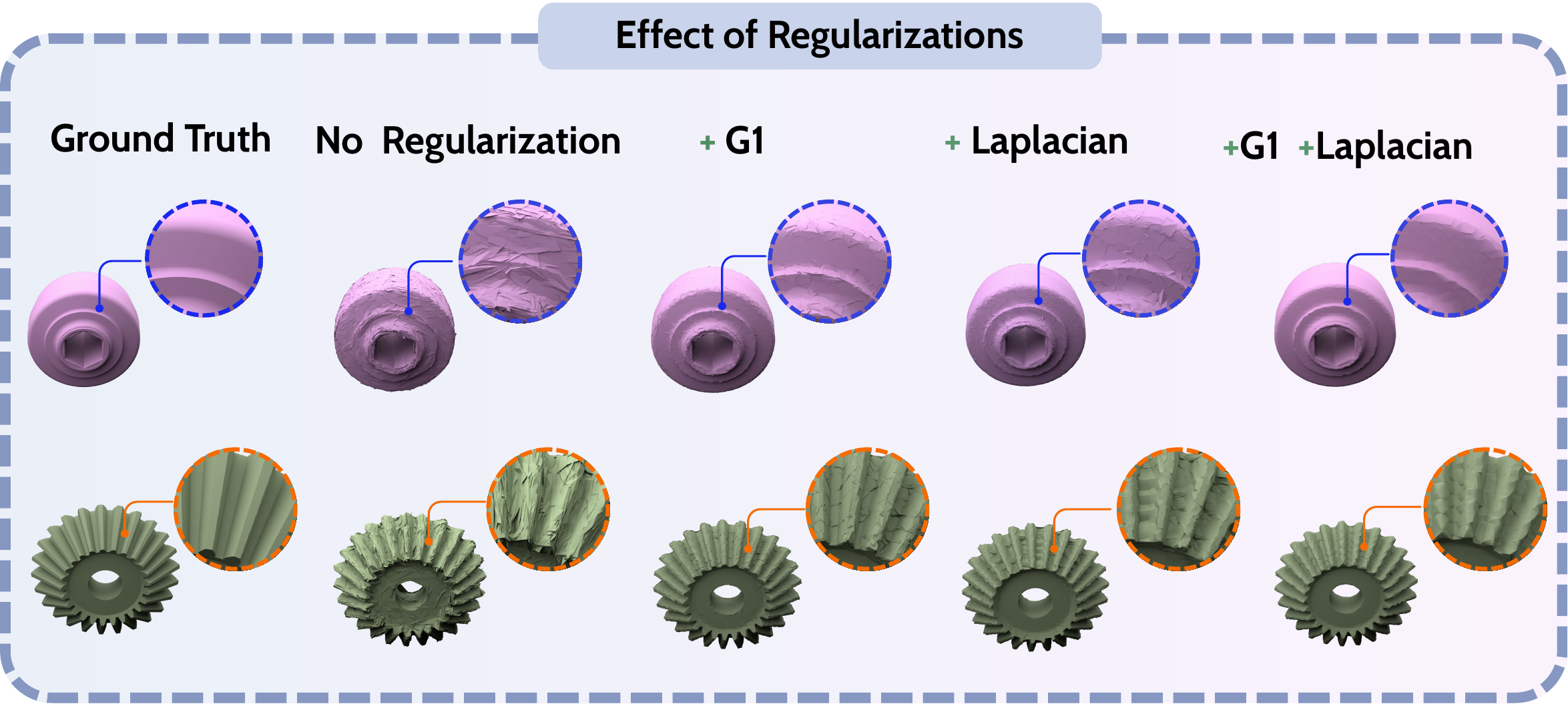}
    \caption{Examples showing VAE reconstructions for different regularizers.}
    \label{fig:regularization}
\end{figure}

\noindent \textbf{Regularizations.} We analyze the impact of \texttt{G1} and \texttt{Laplacian} regularizers on VAE reconstructions. We train the VAE from scratch on 300K samples and evaluating on 15K ABC test shapes (Table~\ref{tab:regularization} and Figure~\ref{fig:regularization}). Without regularization, CD is minimized aggressively but produces rough surfaces with spike artifacts, reflected in the highest Laplacian value. Adding \texttt{G1} or \texttt{Laplacian} individually improves smoothness, with \texttt{Laplacian} better preserving curvature. Combining both gives the lowest Laplacian loss (0.0020) and smoothest surfaces while maintaining strong geometric accuracy. 

\vspace{1mm}
\noindent \textbf{Voxel-Grid Resolution.} We ablate the voxel-grid resolution used to initialize the parametric surfaces (Table~\ref{tab:ablation_combined}.b). Since retraining the VAE at each resolution is expensive, we instead optimize 1000 randomly sampled training meshes. For each resolution, we voxelize each mesh, initialize the parametric surfaces, and optimize for 2000 epochs with learning rate $10^{-4}$ and loss from Eq~\ref{eq:vae_loss}, supervised by ground-truth points. Increasing resolution from $32$ 
to $48$ and $64$ increases patch count by $3\times$ and 
$4\times$, while improving CD by only $1\%$ and $5\%$. 
We therefore adopt resolution $32$ as the best 
quality--efficiency trade-off.

\begin{figure}[h]
    \centering
\includegraphics[width=1\linewidth]{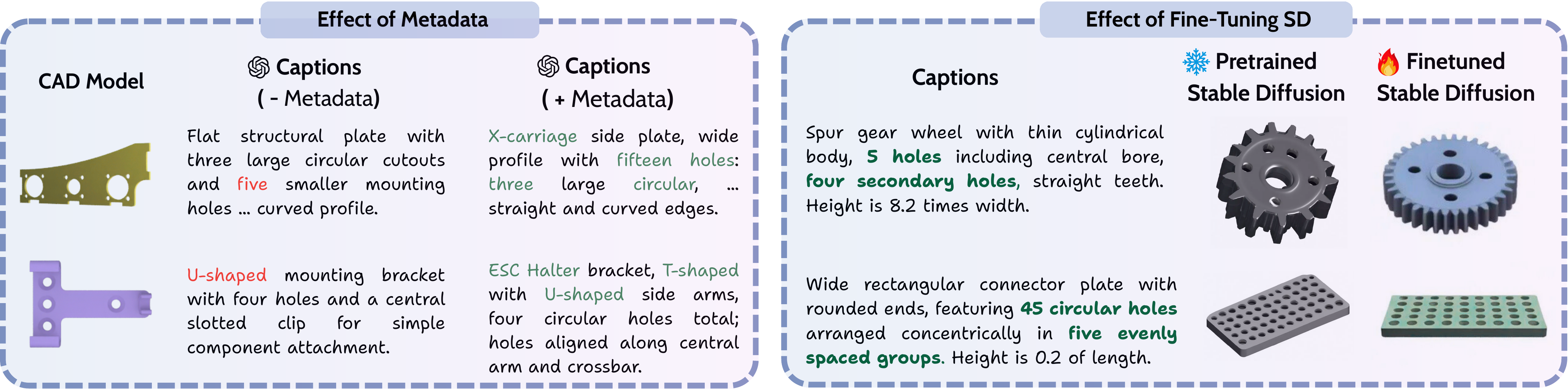}
    \caption{\textbf{Left:} Captions from GPT-5 with and without metadata-augmented prompting. \textbf{Right:} Images from pretrained vs.\ fine-tuned SD-3.5 given the same prompts.}
    \vspace*{-.5\baselineskip}
    \label{fig:metadata-finetuing}
    \vspace*{-.5\baselineskip}
\end{figure}

\vspace{0.1cm}
\noindent \textbf{Metadata Augmentation.} We caption 
$1,000$ random samples with and without metadata using 
GPT-5. Users preferred captions with metadata in 
$80.3$\% of cases, with more accurate part names and 
hole counts (Figure~\ref{fig:metadata-finetuing} Left).


\vspace{0.1cm}
\noindent \textbf{Fine-Tuning Text-to-Image Model.} We compare 500 images generated by pretrained and fine-tuned Stable Diffusion models. User studies show that the fine-tuned model is preferred in $75.6$\% of cases, demonstrating a notable improvement in prompt fidelity (Figure~\ref{fig:metadata-finetuing} -Right).

\vspace{0.1cm}
\noindent\textbf{Flood-fill vs.\ alternative surface extraction.} For VAE-based reconstruction, flood-fill preserves sharp boundaries, which is desirable for CAD, whereas an SDF-based formulation smooths them. It places a heavier burden on the model to recover sharp features. This is reflected in VAE reconstruction error. With both VAEs trained on 100K random samples, flood-fill attains CD $=0.0259$ versus $0.0344$ for SDF. We leave a more extensive study to future work.

\vspace*{-.5\baselineskip}
\begin{table}[ht]
\centering
\vspace*{-.5\baselineskip}
\scriptsize
\vspace*{-.5\baselineskip}
\caption{Ablation studies on initialization quality and coarse-to-fine generation.}
\vspace*{-.5\baselineskip}
\label{tab:supp-ablation}
\setlength{\tabcolsep}{2pt}
\begin{subtable}[t]{0.48\columnwidth}
\centering
\caption{Impact of Initial Parametric Surface.}
\vspace*{-.5\baselineskip}
\label{tab:init-surface}
\resizebox{\columnwidth}{!}{
\begin{tabular}{|ccccc|}
\hline
Noise ($\sigma$) & 0 & $10^{-4}$ & $10^{-3}$ & $10^{-2}$ \\
\hline
CD$\times 10^3$ & 0.034 & 0.039 & 0.041 & 0.065 \\
\hline
\end{tabular}
}
\vspace*{-.5\baselineskip}
\end{subtable}
\hfill
\begin{subtable}[t]{0.48\columnwidth}
\centering
\caption{Importance of Coarse-to-Fine Generation.}
\vspace*{-.5\baselineskip}
\label{tab:coarse-fine}
\resizebox{\columnwidth}{!}{
\begin{tabular}{|ccccc|}
\hline
Noise ($\sigma$) & 0 & $10^{-4}$ & $10^{-3}$ & $10^{-2}$ \\
\hline
CD $\times 10^3$ & 0.93 & 1.01 & 1.23 & 4.51 \\
\hline
\end{tabular}
}
\end{subtable}
\vspace*{-.5\baselineskip}
\vspace*{-.5\baselineskip}
\end{table}

\noindent \textbf{Initial Parametric Surface.} 
We analyze the flood-fill initialization by measuring VAE reconstruction
quality on 15K ABC test samples under Gaussian perturbation of the initial
parametric surface (Table~\ref{tab:init-surface}). The unperturbed init
achieves $CD = 0.034$, confirming a strong geometric prior. The VAE tolerates
small perturbations ($\sigma = 10^{-4}$: $+15\%$ CD) but degrades sharply under
extreme noise ($\sigma = 10^{-2}$: $+91\%$ CD), validating flood-fill as a
robust starting point for parametric surface generation.

\vspace{0.1cm}
\noindent \textbf{Coarse-to-Fine Generation.} 
The VAE decoder generates parametric surfaces from a sparse voxel structure and
per-voxel latents, produced sequentially: coarse voxels first, then
fine-grained latents. To assess the coarse stage, we perturb the voxel grid
with Gaussian noise and evaluate Point2CAD reconstruction on 15K ABC test
samples (Table~\ref{tab:coarse-fine}). The unperturbed structure achieves
$CD = 0.93 \times 10^{-3}$, a strong prior for refinement. The decoder tolerates
small perturbations ($\sigma = 10^{-4}$: $+8.6\%$ CD) but degrades sharply under
extreme noise ($\sigma = 10^{-2}$: $+385\%$ CD), confirming that accurate coarse
geometry is essential and validating our coarse-to-fine design.

\subsection{Application: CAD Topology Recovery} \label{sec:topology}

\begin{figure}[t]
    \centering
\includegraphics[width=1\linewidth]{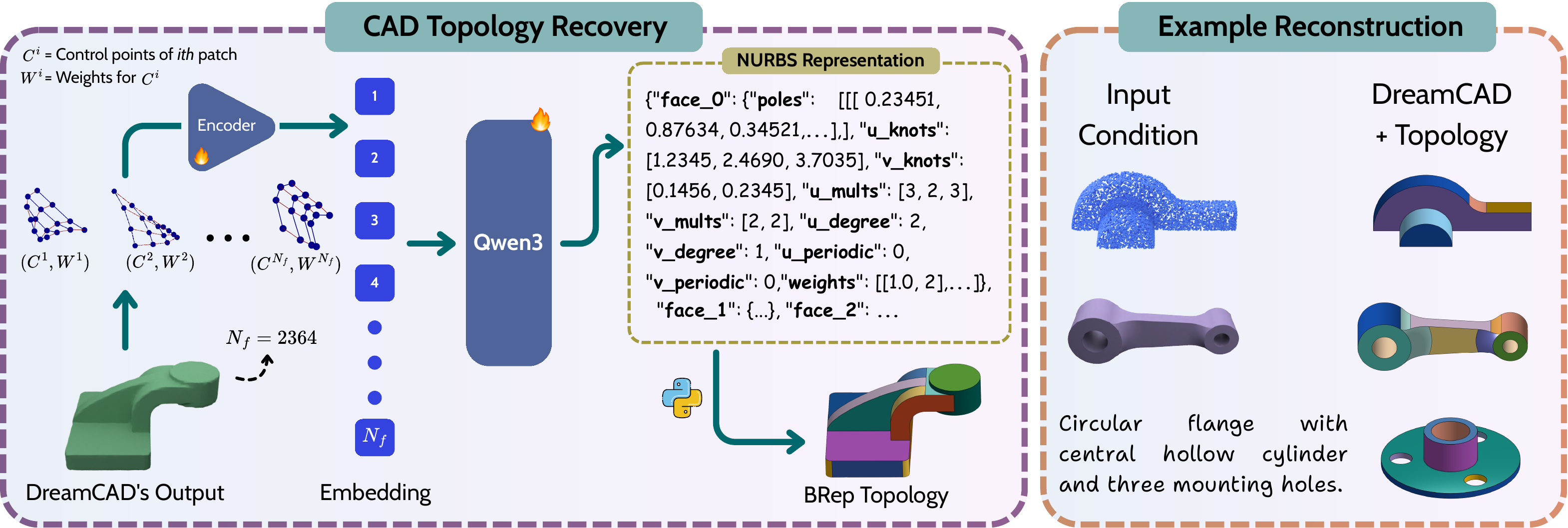}
    \caption{(\textbf{Left}) Topology recovery from patch-based outputs to hybrid NURBS CAD representation~\cite{nurbgen}. (\textbf{Right}) For multimodal inputs, DreamCAD reconstructions and the corresponding recovered topology.}
    \vspace{-.3\baselineskip}
    \label{fig:cadtopology}
     \vspace{-.3\baselineskip}
\end{figure}

DreamCAD's outputs provide control points and weights but lack complete CAD topology - an open challenge we leave to future work. 
As a feasibility study, we fine-tune Qwen3-4B~\cite{qwen3} with LoRA~\cite{lora} on 50K samples to convert patch-based outputs into structured NURBS representations with semantic topology, following NURBGen~\cite{nurbgen}. 
Each patch is represented as 16 control points with 
weights, encoded via a Transformer encoder, and 
passed to Qwen3 for NURBS sequence prediction 
(Figure~\ref{fig:cadtopology} - Left). Since patch counts remain 
low (e.g., $N_f=2364$ in Figure~\ref{fig:cadtopology}), 
training is computationally tractable. Evaluated on 600 
test samples across all three tasks (point-, image-, and 
text-to-CAD, 200 each), this generates 99.2\% valid CAD models (Figure~\ref{fig:cadtopology} - Right) with CD $=0.17\times 10^{-3}$. These results 
suggest that accurate and compact geometric 
reconstruction can provide a strong foundation for 
topology recovery for production-ready CAD generation. Further details are in the supplementary.

\section{Conclusion}


We present \textbf{DreamCAD}, a multi-modal generative framework that produces parametric surfaces directly from point-level supervision. Our goal is to address the long-standing challenge of generalizability in CAD generation. DreamCAD leverages a parametric patch-based CAD representation that supports differentiable mesh generation and enables point-based supervision for 3D shape synthesis. By removing the need for CAD-specific ground-truth annotations, our framework scales to large-scale 3D datasets for CAD geometry generation. In addition, we introduce \textbf{CADCap-1M}, the largest captioning dataset with high-quality textual descriptions generated using GPT-5 for advancing text-to-CAD research. Trained on $1$M+ 3D meshes curated from $10$ public datasets, DreamCAD demonstrates strong generalization across text, image, and point-conditioned generation tasks. While 
complete CAD topology recovery remains a hard and open 
problem, we view DreamCAD as establishing the geometric 
foundation upon which this challenging next stage can 
be built.

\section{Acknowledgement}
This work was in parts supported by the EU Horizon Europe Framework under grant agreement
101135724 (LUMINOUS).
\bibliographystyle{splncs04}
\bibliography{main}

\clearpage


\title{Supplementary of DreamCAD: Scaling Multimodal CAD Generation using Differentiable Parametric Surfaces} 
 \titlerunning{DreamCAD: Scaling Multi-modal CAD Generation}
\author{}
\authorrunning{M.S. Khan et al.}
\institute{}

 \maketitle


\section{More on Data Preparation}

\vspace{1mm}
\noindent \textbf{Visual Feature Generation for Sparse Voxels:} We first normalize each mesh into the range $[-0.5, 0.5]^3$ before voxelization. To generate $150$ multi-view images, we use three complementary camera trajectories that jointly provide full coverage of the object.
(1) \textit{Azimuth sweep:} $50$ images are rendered by rotating the camera around the object with azimuth angles $\theta {=} \{\tfrac{2\pi i}{50}\}_{i=1}^{50}$ and a fixed elevation $\phi {=} 30^\circ$.
(2) \textit{Elevation sweep:} Another $50$ images are captured by keeping the azimuth fixed at $\theta = 30^\circ$ and varying the elevation as $\phi {=} \{\tfrac{2\pi i}{50}\}_{i=1}^{50}$.
(3) \textit{Uniform hemisphere sampling:} The final $50$ views are rendered by uniformly sampling azimuth from $[0, 2\pi]$ and elevation from $[-\tfrac{\pi}{2}, \tfrac{\pi}{2}]$.
All trajectories use a field of view of $40^\circ$, a camera radius of $r {=} 1.5$, and a rendering resolution of $520{\times}520$. Since most meshes in our training dataset lack textures, we assign diffuse colors by randomly sampling RGB values from $[0.5, 0.8]$, resulting in mid-tone to light colors. All $150$ images are rendered using Kaolin’s CUDA-based renderer for faster rendering. During DINO processing, each image is resized to $518{\times}518$ with a patch size of $14$. 

\vspace{1mm}
\noindent \textbf{Filtering Low Quality Models:} As mentioned in Section 5 Experiments section in the main paper, we remove low-quality CAD models from the ABC and Automate datasets. We parse each STEP file using OpenCascade to extract key topological and geometric information, such as the types of surfaces (planes, cylinders, spheres, B-splines, tori, cones, revolutions), the types of edges (lines, circles, B-splines, ellipses), the size of the bounding box, the number of B-Rep faces and vertices, and mean curvature statistics. We first eliminate $99\%$ of the trivial cuboids, which can be reliably detected because they consist of exactly 6 planar faces and 12 straight edges. We also filter out $99\%$ of simple cylindrical objects. These include - (1) basic cylinders with 3 curved surfaces, 1 cylinder and 2 flat end caps (planar surface), (2) very simple cylinder-like shapes with only a single surface and a small number of edges ($<20$), and (3) slightly more complex cylinders that contain only a few cylindrical surfaces ($<5$), few edges ($<20$) and no torus, cone, or revolved surfaces. We further discard degenerate or physically unrealistic models by removing objects with excessively large bounding boxes or abnormally high curvature values. Also models with lower number of faces ($<5$), or vertices ($<10$), or edges ($<10$) are removed.

\vspace{1mm}
\noindent \textbf{Generation Dataset Preparation:} 
For data preparation stage during \textit{image-to-CAD} training, we render 4 images per CAD shape following the same camera setup as MARVEL~\cite{marvel}. 
Only for textureless models, we apply random diffuse colors selected from a curated 30+ dark color palette, which includes 
\colname{deepnavy}{deep navy}, 
\colname{charcoal}{charcoal}, 
\colname{deepplum}{plum}, 
\colname{mossgreen}{moss green}, 
\colname{stormblue}{storm blue}, 
\colname{aubergine}{aubergine}, 
\colname{mulberry}{mulberry}, 
and other low-saturation tones.
We use these images during finetuning of Stable-Diffusion 3.5 for \textit{text-to-image} task as well.


\begin{figure}[ht]
    \centering
    \includegraphics[width=0.75\linewidth]{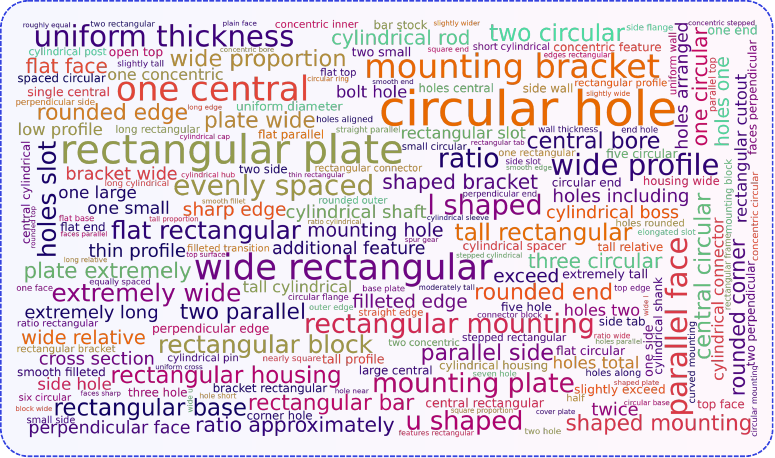}
    \caption{Wordcloud of captions from CADCap-1M}
    \vspace*{-.3\baselineskip}
    \label{fig:wordcloud}
    \vspace*{-.3\baselineskip}
\end{figure}

\vspace*{-.6\baselineskip}
\section{More on CADCap-1M}

\noindent For GPT-5–based captioning, we processed over 1M samples using the batch API (batch size 2k) over three weeks at a total cost of $\$1800$. For metadata augmentation, approximately 20\%, 39\%, and 46\% of samples in ABC, Automate, and Fusion360 contain part names.

\vspace{1mm}
\noindent In Table~\ref{tab:caption}, we provide summary statistics of the CADCap-1M captions. The mean caption length is under 20 words. The vocabulary diversity is substantial, with over $21$k unigrams, $446$k bigrams, and $2.3$M trigrams, indicating rich linguistic variation for robust \textit{text-to-CAD} learning. In Figure~\ref{fig:wordcloud}, we provide the wordcloud of the captions.

\begin{table}[ht]
\def\arraystretch{1.3}%
\caption{Summary statistics of the CADCap-1M dataset, including average caption length and diversity measured through unigram, bigram, and trigram counts.}
\label{tab:caption}
\centering
\resizebox{0.6\columnwidth}{!}{
\begin{tabular}{cccc}
\hline
\cellcolor{blue!10} Mean Length & \cellcolor{blue!10} Unigram & \cellcolor{blue!10} Bigram & \cellcolor{blue!10} Trigram \\ \hline
19.63 & 21042   & 446,061 & 2,368,731 \\ \hline
\vspace*{-.5\baselineskip}
\end{tabular}
}
\vspace*{-.5\baselineskip}

\end{table}

\begin{figure*}[ht]
     \centering
    \includegraphics[width=0.95\linewidth]{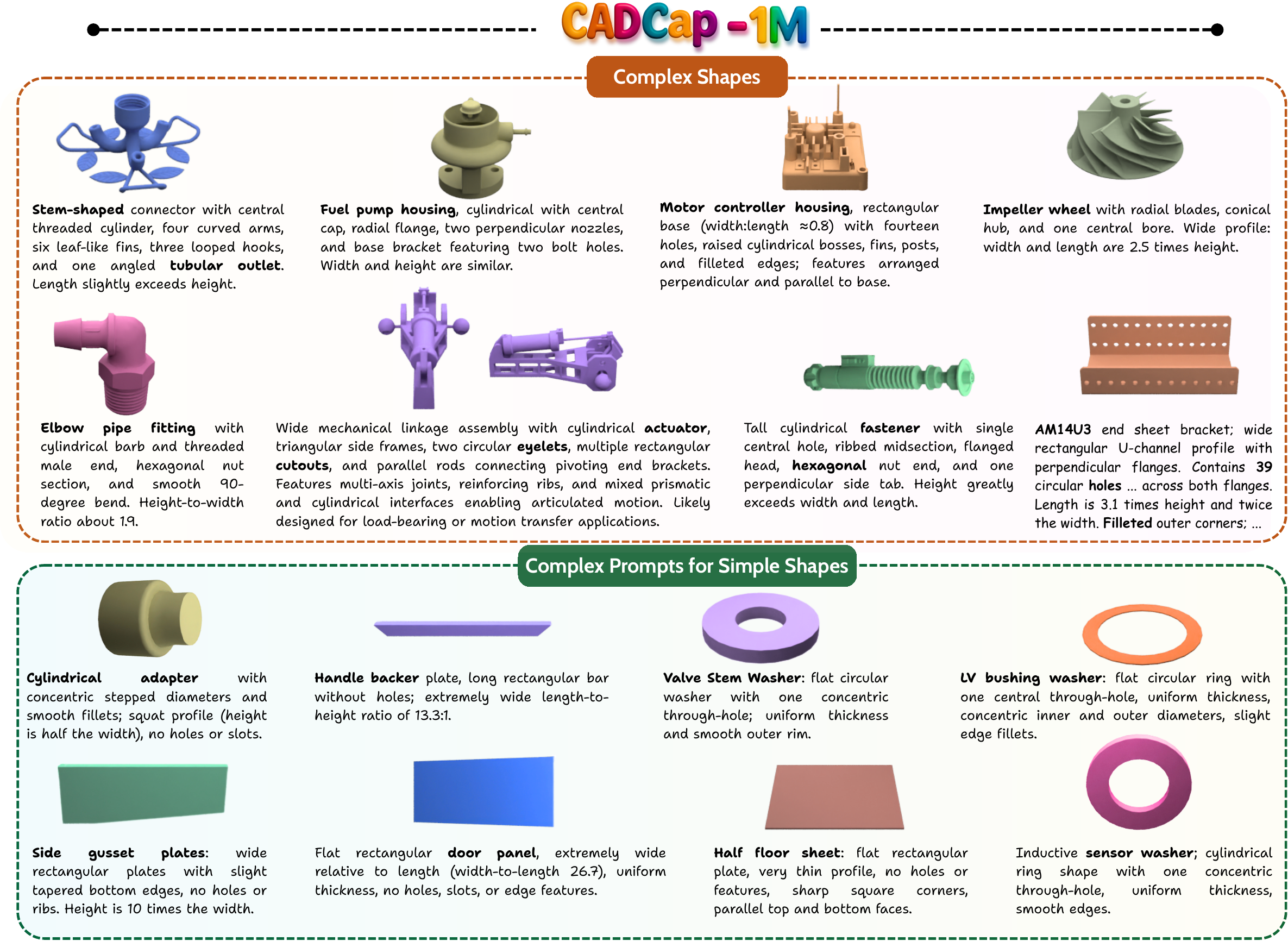}
    \caption{Examples of captions for \textbf{complex} (top) and \textbf{simple} (bottom) CAD parts.}
    \label{fig:caption_1}
\end{figure*}

\begin{figure*}[t]
     \centering
    \includegraphics[width=0.95\linewidth]{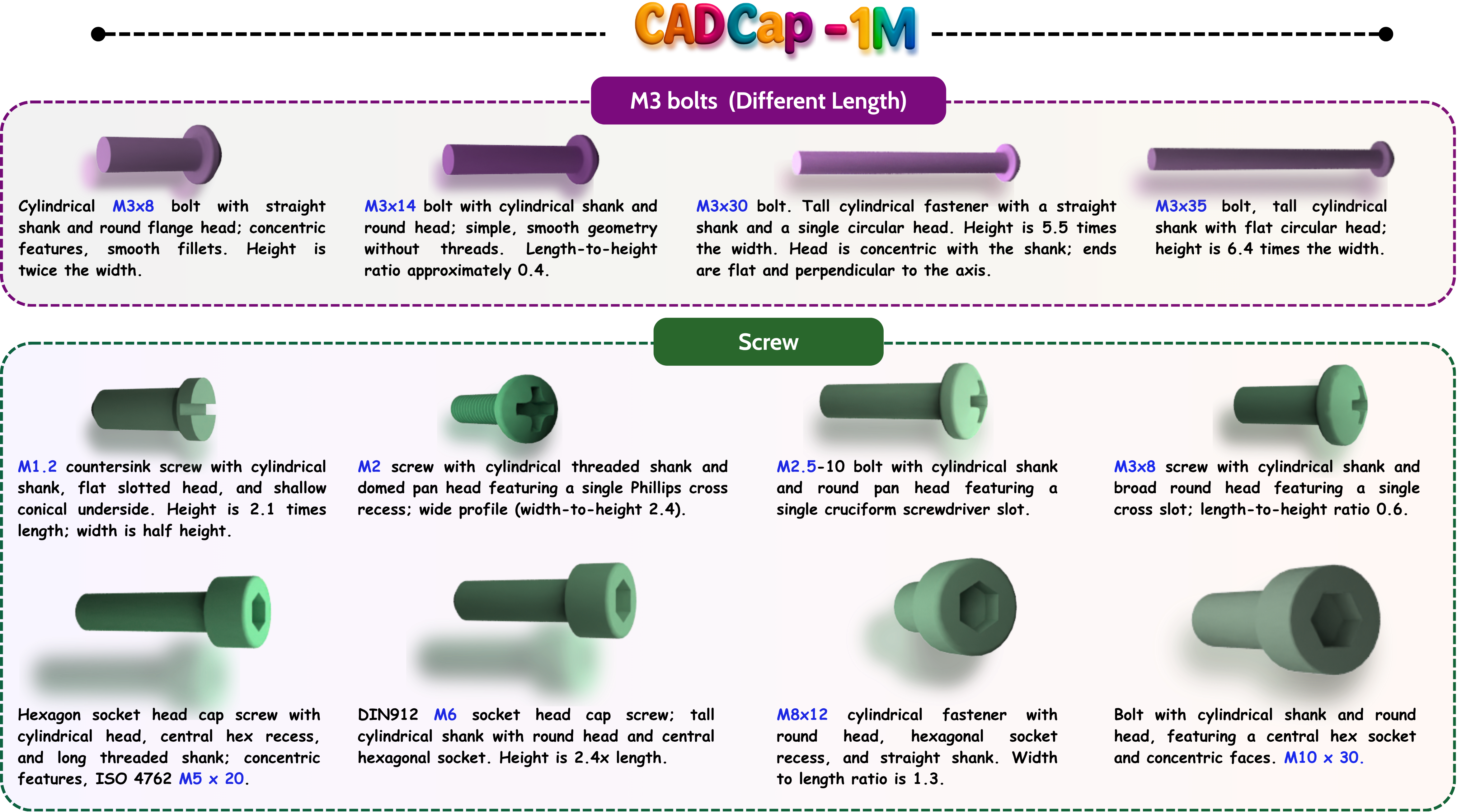}
   \caption{Examples of captions for different types of \textbf{fasteners}.}
    \label{fig:caption_3}
\end{figure*}

\begin{figure*}[ht]
     \centering
    \includegraphics[width=1\linewidth]{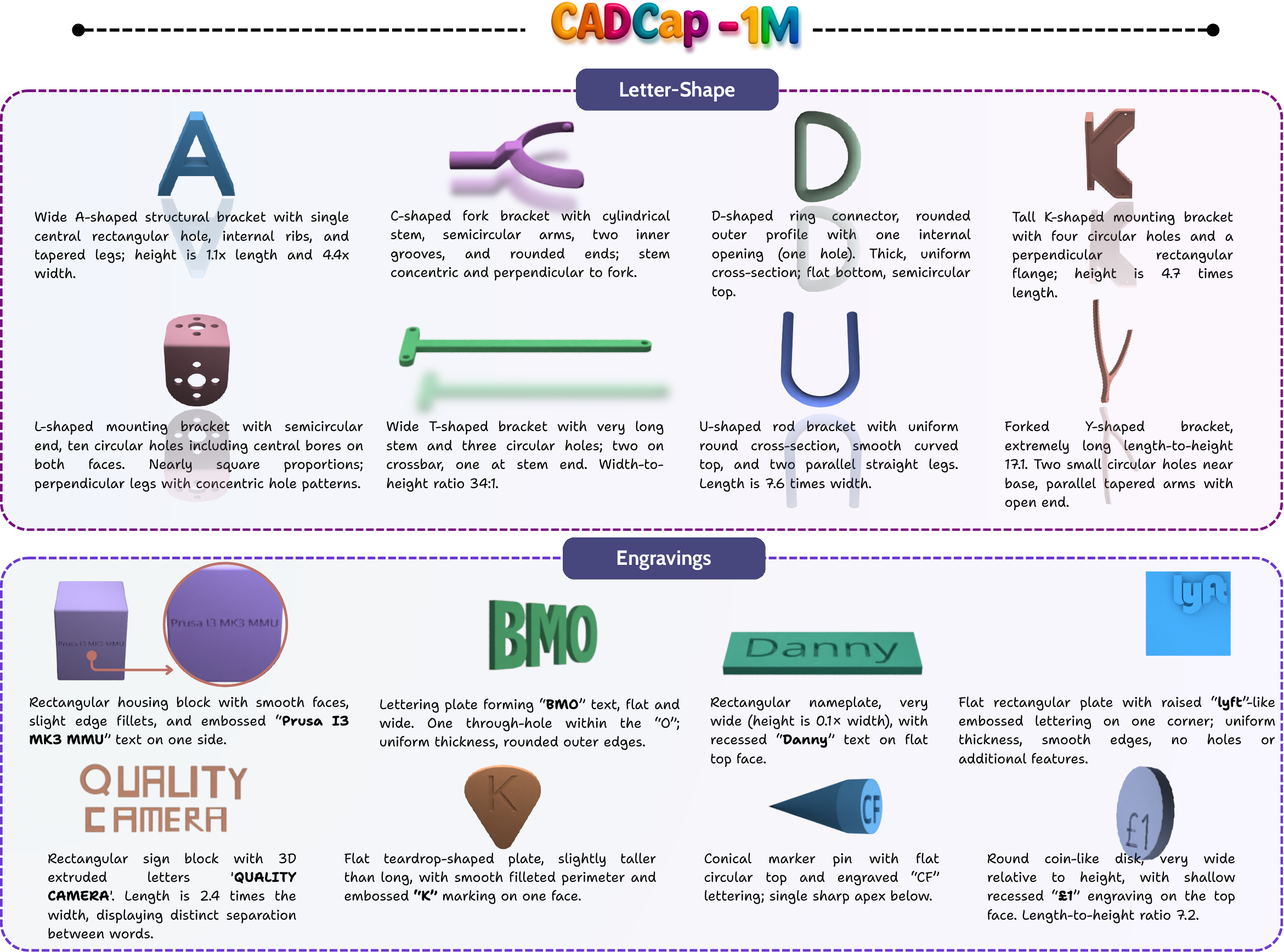}
    \caption{Examples of captions for \textbf{letter}-based (top) and \textbf{engraving}-style (bottom) CAD models.}
    \label{fig:caption_4}
\end{figure*}

\vspace{1mm}
\noindent In Figures~\ref{fig:caption_1},~\ref{fig:caption_3}, and~\ref{fig:caption_4}, we present representative examples of the generated captions. In Figure~\ref{fig:caption_1} (top), captions for complex shapes are shown, where our pipeline produces highly detailed descriptions such as “\textit{Stem-shaped connector … six leaf-like fins}” or “\textit{AM14U3 end sheet … 39 circular holes}.” In Figure~\ref{fig:caption_1} (bottom), we highlight the effect of metadata inclusion, especially part names, which significantly improves the specificity of captions for simpler shapes. For example, cases 3, 4, and 8 demonstrate that visually similar ring-like structures are correctly identified as “\textit{Valve Stem Washer},” “\textit{LV Bushing Washer},” and “\textit{Inductive Sensor Washer},” respectively. Such fine-grained captions are essential because many small components in larger assemblies are geometrically simple. In Figure~\ref{fig:caption_3}, we present captions that distinguish between different screw and bolt types (e.g., \textit{M1.2}, \textit{M2}, \textit{M10}) as well as variations in length for the same bolt type (e.g., \textit{M3×8}, \textit{M3×14}). Finally, Figure~\ref{fig:caption_4} illustrates captions generated for letter-based or engraving-style CAD models.


\begin{figure}[t]
    \centering
    \includegraphics[width=0.7\linewidth]{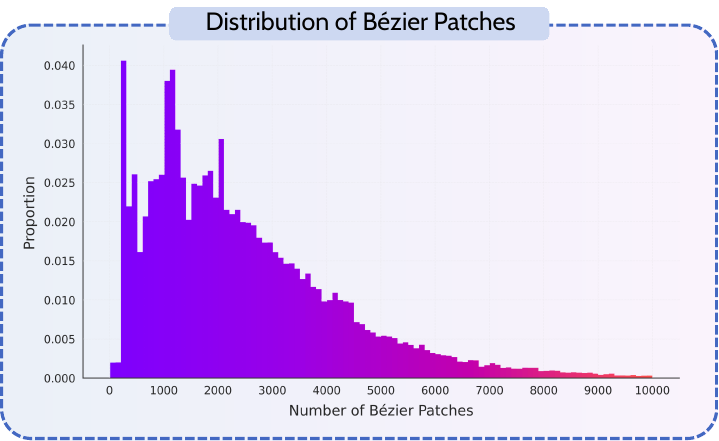}
    \caption{Distribution of patch sizes in training datasets.}
    \vspace*{-.5\baselineskip}
    \label{fig:patch_size}
    \vspace*{-.5\baselineskip}
\end{figure}

\vspace{-.5\baselineskip}
\section{More Experimental Results}
\vspace{-.5\baselineskip}

\noindent \textbf{Training Details:} For VAE training, we adopt the same weight initialization scheme and KL weighting as in Trellis~\cite{trellis}. The output layer predicting the deformation vector is initialized with $N(0,10^{-5})$, and the control-point weight prediction layer is zero-initialized with its bias set to 1. This initialization encourages the model to produce only minimal deformations in the early stages, substantially reducing the likelihood of unstable spikes or surface artifacts compared to random initialization. Since DreamCAD handles CAD models with widely varying geometric complexity, the VAE is trained on a diverse set of patch configurations. Figure~\ref{fig:patch_size} shows the distribution of Bézier patch counts across training samples.

\begin{table}[ht]
\caption{Parameter counts for all components of DreamCAD.}
\vspace{-.7\baselineskip}
\label{tab:parameter}
\centering
\def\arraystretch{1.3}
\resizebox{0.4\columnwidth}{!}{
\begin{tabular}{c:c:cc:cc}
\hline
\multirow{2}{*}{\textbf{VAE}} & \multirow{2}{*}{\textbf{\begin{tabular}[c]{@{}c@{}}Sparse \\ Structure\end{tabular}}} & \multicolumn{2}{c:}{\textbf{\begin{tabular}[c]{@{}c@{}}Coarse\end{tabular}}} & \multicolumn{2}{c}{\textbf{Fine-Grained}} \\
                              &                                                                                       & \textbf{Image}                                                      & \textbf{Point}                                   & \textbf{Image} & \multicolumn{1}{c}{\textbf{Point}} \\ \hline
71M                           & \multicolumn{1}{c:}{133M}                                                                  & 354M                                   & 280M                                    & 400M  & 325M                      \\ \hline
\end{tabular}
}

\vspace*{-.6\baselineskip}
\end{table}

\begin{table}[t]
\def\arraystretch{1.2}%
\caption{Quantitative comparison on Point2CAD, Img2CAD, and Text2CAD tasks over the DeepCAD test set. F1 is scaled by $10^2$, while CD, JSD, and MMD by $10^3$. For \textit{text}- and \textit{image-to-CAD}, GPT and User ratings measure visual alignment.}
\label{tab:deepcad-quan}
\centering
\setlength{\tabcolsep}{2.5pt}
\resizebox{0.65\columnwidth}{!}{
\begin{tabular}{ll:cccccc}
\hline
\multicolumn{1}{c}{\multirow{2}{*}{Task}} & \multicolumn{1}{c}{\multirow{2}{*}{Models}} 
& \multicolumn{6}{c}{DeepCAD} \\ 
\multicolumn{1}{c}{} & \multicolumn{1}{c}{} 
& \cellcolor{ubc!20}F1 $\uparrow$ & \cellcolor{ubc!20}NC $\uparrow$ & \cellcolor{blue!10}CD $\downarrow$ & \cellcolor{blue!10}HD $\downarrow$ & \cellcolor{blue!10}JSD $\downarrow$ & \cellcolor{blue!10}MMD $\downarrow$  \\ \hline
\addlinespace[2pt]
\multirow{4}{*}{\rotatebox{90}{\textbf{Point}}} 
& DeepCAD & 23.18 & 0.53 & 47.63 & 0.34 & 729.11 & 26.80 \\
& CAD-Recode & 90.95 & 0.92 & 1.38 & 0.06 & 126.15 & 1.13 \\
& Cadrille & 92.12 & 0.93 & 0.30 & 0.05 & 112.31 & 1.05 \\
& \textbf{DreamCAD} & \textbf{94.91} & \textbf{0.94} & \textbf{0.12} & \textbf{0.03} & \textbf{98.51} & \textbf{0.90} \\ \hline
\addlinespace[1pt]
\multirow{3}{*}{\rotatebox{90}{\textbf{Img}}} 
& & \multicolumn{1}{c}{\cellcolor{ubc!20}GPT $\uparrow$} & \multicolumn{1}{c}{\cellcolor{ubc!20}User $\uparrow$} & \multicolumn{1}{c}{\cellcolor{blue!10}CD $\downarrow$} & \multicolumn{1}{c}{\cellcolor{blue!10}HD $\downarrow$} & \multicolumn{1}{c}{\cellcolor{blue!10}JSD $\downarrow$} & \multicolumn{1}{c}{\cellcolor{blue!10}MMD $\downarrow$} \\ 
& Cadrille & 2.10 & 3.30 & 78.13 & 0.46 & 856.31 & 36.13 \\
& \textbf{DreamCAD} & \textbf{97.90} & \textbf{96.7} & \textbf{12.41} & \textbf{0.20} & \textbf{641.35} & \textbf{26.13} \\ \hline
\addlinespace[1pt]
\multirow{9}{*}{\rotatebox{90}{\textbf{Text}}} 
& & \multicolumn{1}{c}{\cellcolor{ubc!20}GPT $\uparrow$} & \multicolumn{1}{c}{\cellcolor{ubc!20}User $\uparrow$} & \multicolumn{1}{c}{\cellcolor{blue!10}CD $\downarrow$} & \multicolumn{1}{c}{\cellcolor{blue!10}HD $\downarrow$} & \multicolumn{1}{c}{\cellcolor{blue!10}JSD $\downarrow$} & \multicolumn{1}{c}{\cellcolor{blue!10}MMD $\downarrow$} \\ 
& DeepCAD & 0.90 & 1.89 & 80.56 & 0.39 & 813.45 & 36.25 \\
& Text2CAD & 10.3 & 19.87 & 82.77 & 0.39 & 759.86 & 34.91 \\
& Cadrille & 2.28 & 3.58 & 168.86 & 0.56 & 958.22 & 100.94 \\
& T2CQ (Q3B) & 0.00 & 0.00 & 84.32 & 0.39 & 859.51 & 38.68 \\
& T2CQ (G2L) & 0.01 & 0.00 & 87.47 & 0.38 & 851.30 & 38.34 \\
& T2CQ (CG) & 0.34 & 0.35 & 77.78 & 0.38 & 860.21 & 38.50 \\
& CADFusion & 15.98 & 14.02 & 38.92 & 0.26 & 751.54 & 32.16 \\
& \textbf{DreamCAD} & \textbf{70.19} & \textbf{74.31} & \textbf{23.11} & \textbf{0.16} & \textbf{741.19} & \textbf{28.56} \\ \hline   
\end{tabular}
}
\vspace*{-.5\baselineskip}
\end{table}


\vspace{1mm}
\noindent For both the coarse and fine-grained Flow Transformers, we use 12 Transformer decoder layers. The flow-matching models are trained using a logit-normal time schedule with $\mu{=}0$ and $\sigma{=}1$, and optimized with AdamW using a learning rate of $5\times10^{-5}$.  Table~\ref{tab:parameter} lists DreamCAD's module-wise parameter counts.




\vspace{1mm}
\noindent \textbf{Comparison on DeepCAD Test Set:} To isolate our architectural contributions from data scale, we train DreamCAD from scratch on the DeepCAD dataset using identical settings, comparing only against baselines trained on the same data. As shown in Table~\ref{tab:deepcad-quan}, DreamCAD outperforms all baselines across \textit{point}-, \textit{image}-, and \textit{text-to-CAD} tasks with zero invalid outputs, demonstrating that our multi-stage geometry-based learning captures shape structure more effectively than sequential token-based approaches, even at limited data scales.

\begin{figure}[t]
    \centering
    \includegraphics[width=0.6\linewidth]{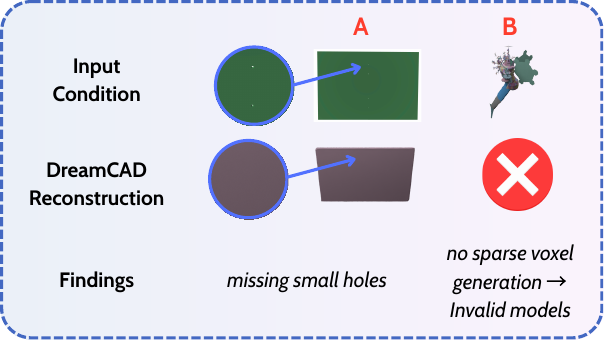}
    \caption{Some failure cases showcasing missing geometric details and invalid models.}
    \vspace*{-.5\baselineskip}
    \label{fig:failure}
    \vspace*{-.5\baselineskip}
\end{figure}

\vspace{1mm}
\noindent \textbf{Failure Cases:} Figure~\ref{fig:failure} illustrates two failure modes of DreamCAD. First, fine geometric details such as tiny holes can be lost when the sparse voxel resolution is insufficient to capture them. This can be addressed in future work by adopting octree-based representations for finer geometric capture. Second, ambiguous images occasionally fail to generate any active voxels in the coarse stage, resulting in invalid outputs.

\begin{figure*}[t]
     \centering
    \includegraphics[width=1\linewidth]{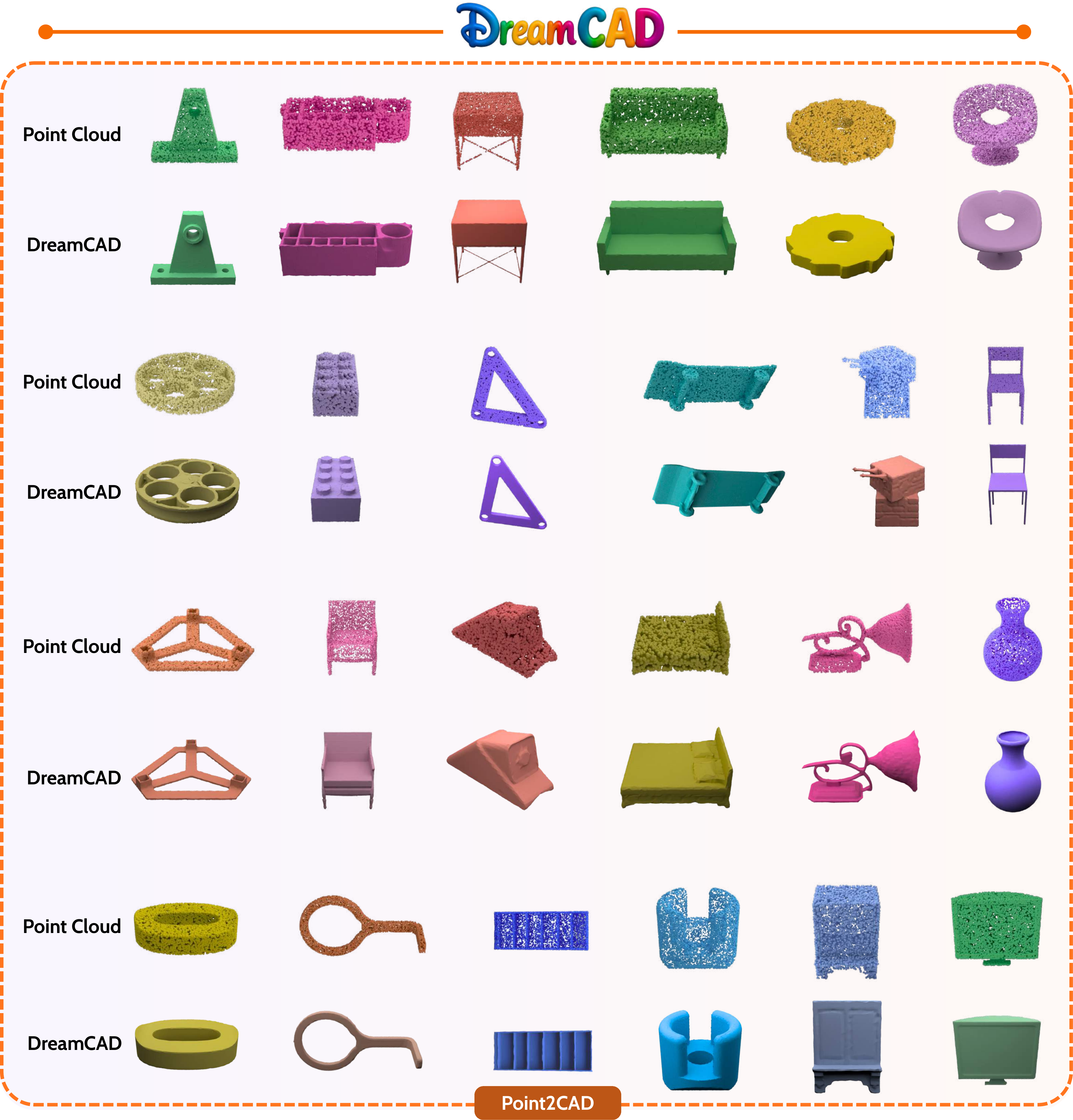}
    \caption{More qualitative results for the \textbf{\textit{point-to-CAD}} reconstruction task using DreamCAD. Each row shows input point clouds (top) and the corresponding reconstructed CAD models (bottom) generated by DreamCAD across a wide range of shapes, including mechanical components, furniture, utensils, and free-form surfaces. As seen in the examples, DreamCAD successfully recovers clean parametric geometry from input point clouds, preserving fine structural details (e.g., \textit{tubular connectors, circular cutouts, brackets}), smooth surfaces (e.g., \textit{bowls, vases, seats}), and complex topologies (e.g., \textit{multi-part assemblies and chair frames}).}
    \label{fig:point2cad}
\end{figure*}

\begin{figure*}[t]
     \centering
    \includegraphics[width=1\linewidth]{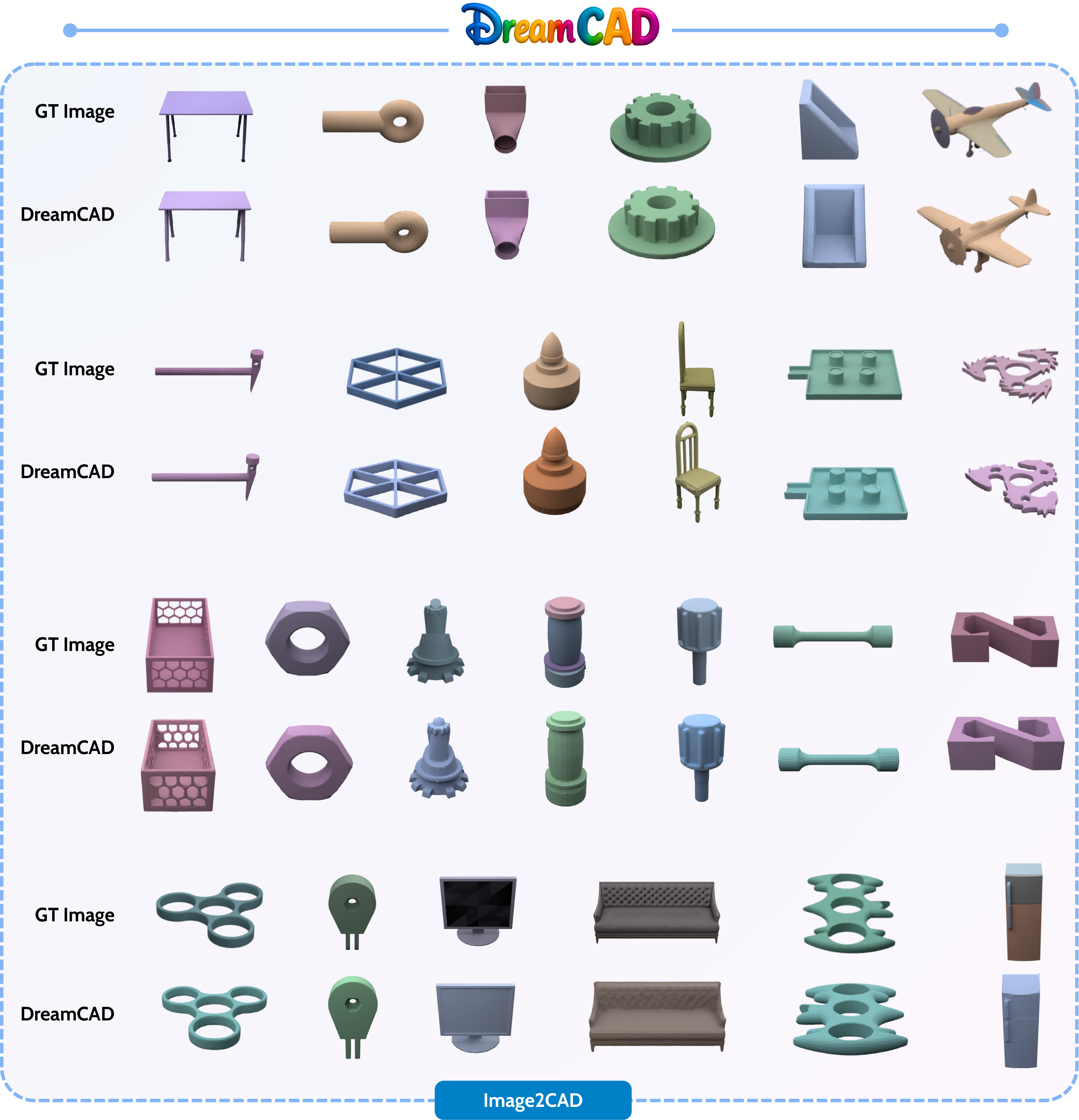}
    \caption{More qualitative results for the \textit{\textbf{image-to-CAD}} generation task using DreamCAD. Each row shows the ground-truth reference image (top) and the CAD reconstruction produced by DreamCAD (bottom) across a wide range of object categories, including furniture, mechanical parts, consumer products, and free-form designs. As seen in the examples, DreamCAD consistently recovers accurate 3D geometry from a single image, capturing fine structural details (e.g., \textit{chair backrests, table legs, nozzle openings}), complex mechanical features (e.g., \textit{gear teeth, connectors, threads}), and overall proportions with high fidelity.}
    \label{fig:img2cad}
\end{figure*}

\begin{figure*}[t]
     \centering
    \includegraphics[width=1\linewidth]{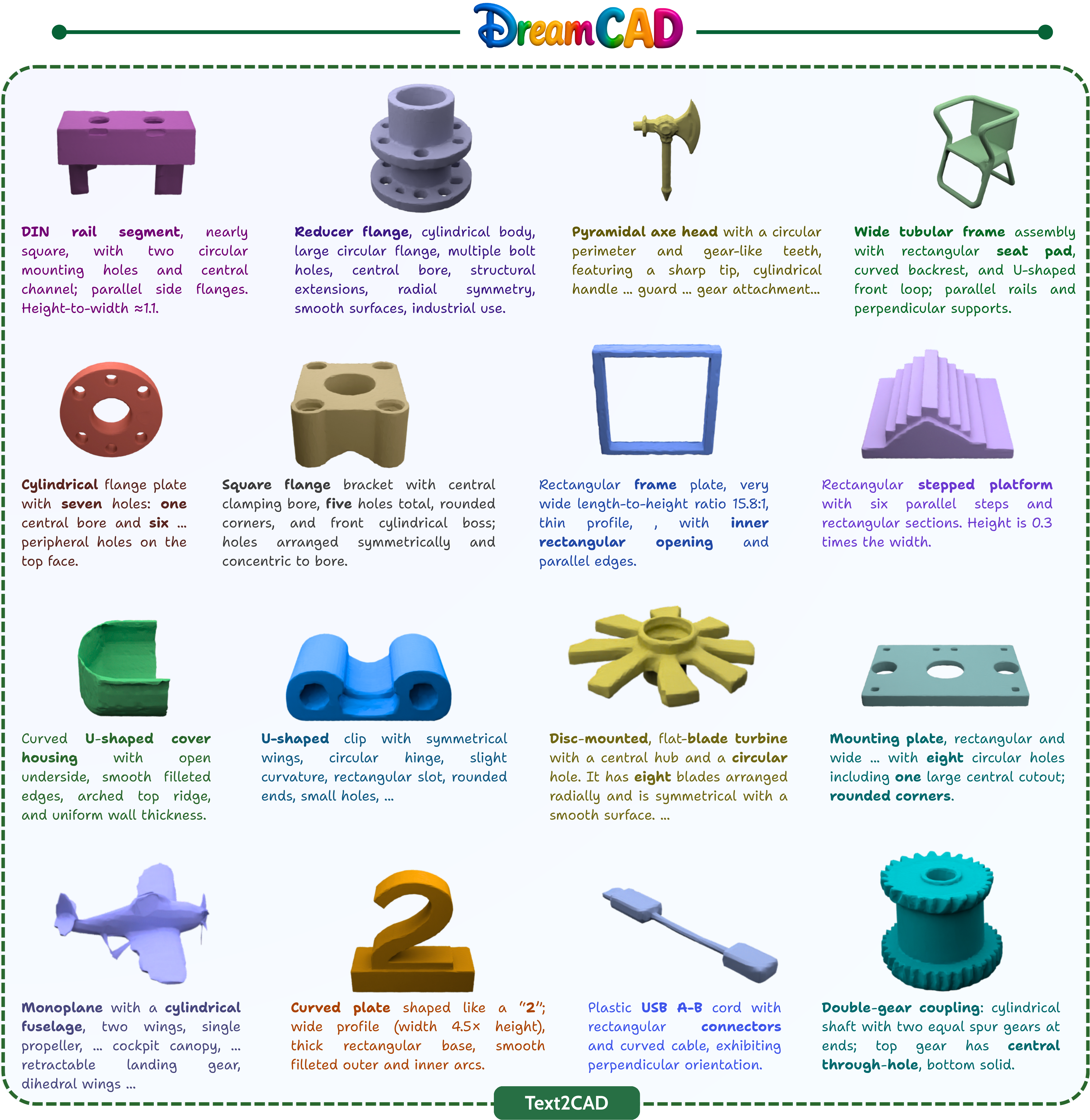}
    \caption{More qualitative \textit{ \textbf{text-to-CAD} } results on the test set. Each row shows an input text description (bottom) and the corresponding CAD geometry generated by DreamCAD (top). DreamCAD successfully reconstructs shapes from complex text prompts, ranging from mechanical parts (\textit{gear assemblies, flange plates, couplings}), tools (\textit{pyramidal axe head}), and structural elements (\textit{DIN rail, stepped platforms}), to free-form designs (\textit{seat frames, turbine blades}), and consumer objects (\textit{USB cable}).}
    \label{fig:text2cad_1}
\end{figure*}

\begin{figure*}[t]
     \centering
    \includegraphics[width=1\linewidth]{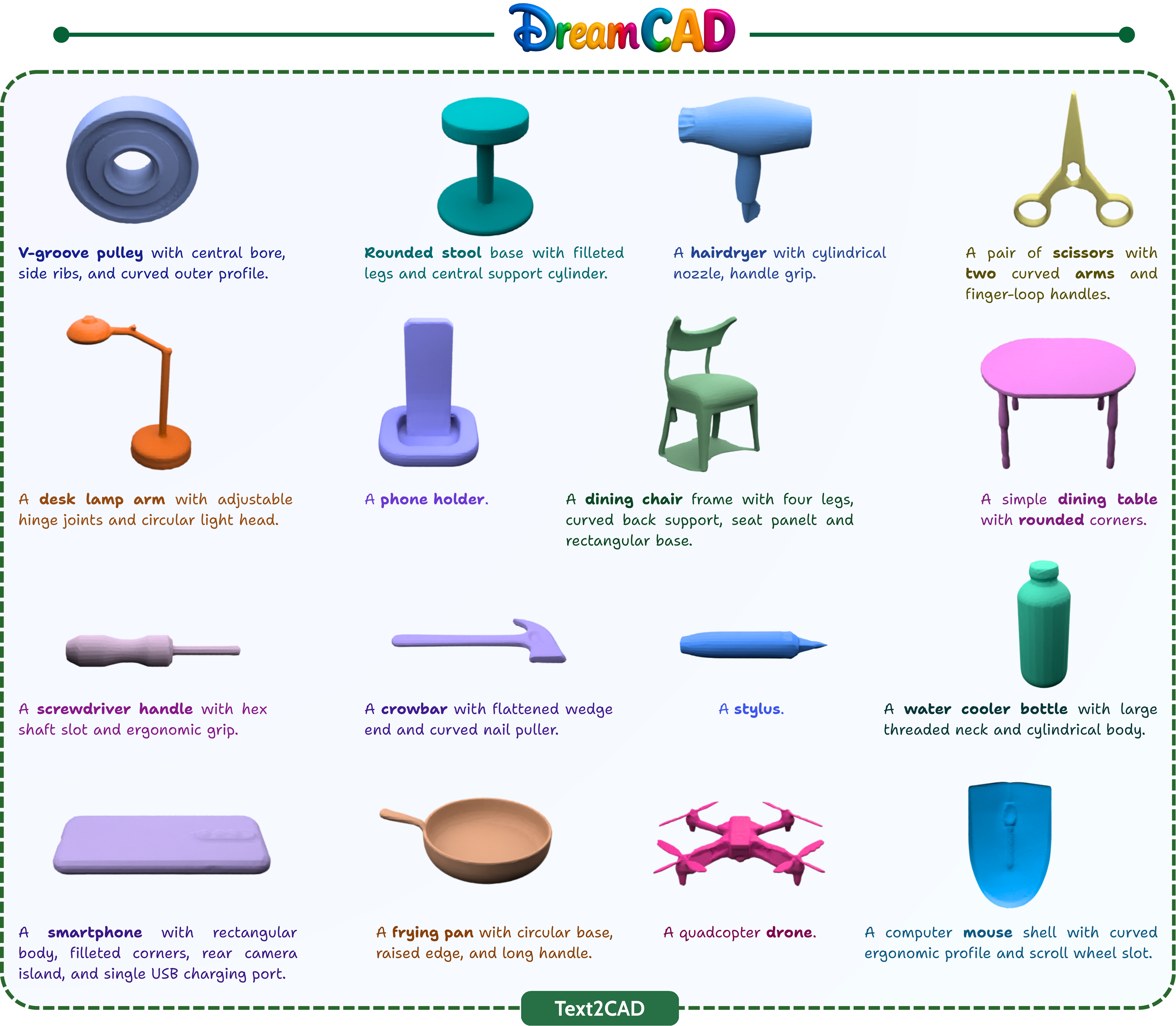}
    \caption{\textit{\textbf{Text-to-CAD}} generation results on GPT-generated prompts. Each example shows a text prompt produced by GPT-5 (bottom) and the corresponding CAD reconstruction (top) by DreamCAD. As shown, DreamCAD generalizes beyond dataset-style prompts and reliably interprets free-form, open-vocabulary instructions which includes consumer electronics (\textit{smartphone}, \textit{computer mouse}), household items (\textit{frying}\textit{ pan, stool, water cooler bottle}), tools (\textit{screwdriver, crowbar, scissors}), mechanical components (\textit{V-groove pulley}), and articulated objects (\textit{desk lamp arm, quadcopter drone}).}
    \label{fig:text2cad_2}
\end{figure*}

\vspace{1mm}
\noindent \textbf{More Conditional Generation Results:} Figure~\ref{fig:point2cad},~\ref{fig:img2cad},~\ref{fig:text2cad_1},~\ref{fig:text2cad_2} provide additional qualitative results for \textit{point}-, \textit{image}-, and \textit{text-to-CAD}.

\vspace{1mm}
\noindent \textbf{More Details on Topology Recovery:} As discussed in
Section~\ref{sec:topology}, we fine-tune Qwen3-4B~\cite{qwen3} to generate
symbolic NURBS representations from DreamCAD's patch-based outputs. We sample
50K BReps from the ABC dataset. We use OpenCascade to generate the hybrid NURBS-based representation~\cite{nurbgen} of the BReps. We retain only those samples whose NURBS representation contains fewer than 10K tokens. To generate the input Bézier surfaces for each BRep, we first convert it to a mesh, voxelize it in $32^3$ grid-resolution. Then we generate the initial parametric surface, and run optimization for 2000 epochs with learning rate $10^{-4}$ using the loss from Eq.~\ref{eq:vae_loss}. The resulting patch-based representations are used to fine-tune Qwen3-4B following~\cite{nurbgen} for 3 epochs with a context window of 20K tokens.

\section{More Ablation Studies}

\begin{table}[ht]
\centering
\scriptsize
\caption{Ablation studies on initialization quality and coarse-to-fine generation.}
\label{tab:supp-ablation}
\setlength{\tabcolsep}{2pt}
\begin{subtable}[t]{0.48\columnwidth}
\centering
\caption{Impact of Initial Parametric Surface.}
\label{tab:init-surface}
\resizebox{\columnwidth}{!}{
\begin{tabular}{|ccccc|}
\hline
Noise ($\sigma$) & 0 & $10^{-4}$ & $10^{-3}$ & $10^{-2}$ \\
\hline
CD$\times 10^3$ & 0.034 & 0.039 & 0.041 & 0.065 \\
\hline
\end{tabular}
}
\end{subtable}
\hfill
\begin{subtable}[t]{0.48\columnwidth}
\centering
\caption{Importance of Coarse-to-Fine Generation.}
\label{tab:coarse-fine}
\resizebox{\columnwidth}{!}{
\begin{tabular}{|ccccc|}
\hline
Noise ($\sigma$) & 0 & $10^{-4}$ & $10^{-3}$ & $10^{-2}$ \\
\hline
CD $\times 10^3$ & 0.93 & 1.01 & 1.23 & 4.51 \\
\hline
\end{tabular}
}
\end{subtable}
\vspace*{-.5\baselineskip}
\vspace*{-.5\baselineskip}
\end{table}

\noindent \textbf{Impact of Initial Parametric Surface.} 
We analyze the role of the flood-fill initialization 
by evaluating VAE reconstruction quality on 15K ABC 
test samples under varying levels of Gaussian 
perturbation applied to the initial parametric surface. 
As shown in Table~\ref{tab:init-surface}, the 
unperturbed initialization achieves $CD = 0.034$, 
confirming that flood-fill provides a strong geometric 
prior. The VAE remains robust to small perturbations 
($\sigma = 10^{-4}$: 15\% CD increase), demonstrating 
that the decoder can recover from minor initialization 
errors. However, performance degrades significantly 
under extreme noise ($\sigma = 10^{-2}$: 91\% CD 
increase). These results validate our choice of flood-fill 
initialization as an effective and robust starting 
point for parametric surface generation.

\vspace{0.1cm}
\noindent \textbf{Importance of Coarse-to-Fine Generation.} 
The VAE decoder requires both a sparse voxel structure 
and per-voxel latent features to generate parametric 
surfaces, produced sequentially: coarse voxels first, 
then fine-grained latents. To assess the importance of the coarse stage, we perturb the voxel grid with 
Gaussian noise and evaluate the Point-to-CAD reconstruction 
on 15K ABC test samples. As shown in the Table~\ref{tab:coarse-fine}, the 
unperturbed coarse structure achieves $CD = 0.93 \times 
10^{-3}$, serving as a strong geometric prior for the 
refinement stage. The decoder remains robust to small 
perturbations ($\sigma = 10^{-4}$: 8.6\% CD increase), 
but degrades significantly under extreme noise 
($\sigma = 10^{-2}$: 385\% CD increase). These results 
confirm that accurate coarse geometry is essential for 
high-quality parametric surface generation, validating 
our coarse-to-fine design choice.

\section{Disscussion on Future Research Directions} \label{sec:supp-application}

DreamCAD addresses the first and one of most challenging stages of scalable CAD generation: learning accurate geometric reconstruction from large-scale unstructured 3D data without CAD-specific annotations. A natural question is whether \textit{patch-based parametric surfaces are sufficient for real industrial workflows}. We argue that patch-based surfaces are not the final goal, but rather a necessary and tractable foundation.

\vspace{0.1cm}
\noindent Our key insight is that while recovering CAD topology remains a challenging 
problem in its own right, jointly generating geometry 
and topology from multimodal inputs at scale is 
a harder problem. The former can 
leverage decades of reverse engineering 
research~\cite{complexgen,brepdetnet}, well-defined 
geometric primitives~\cite{nurbgen}, and existing fitting 
tools~\cite{parsenet,point2cad}. The latter requires simultaneous 
reasoning over noisy inputs, diverse 3D geometries, and 
complex topological structure, with no comparable 
infrastructure to draw upon. As a first step 
toward topology recovery, our experiment in 
Section~\ref{sec:topology} DreamCAD's outputs can be converted into production-level CAD models.


\vspace{0.1cm}
\noindent Future work should build on DreamCAD's outputs toward full CAD topology recovery. Promising directions include NURBS-face reconstruction via patch-merging, sharp feature detection and edge topology inference, hierarchical part-based editing, enforcing higher-order continuity ($C^1$/$G^1$), and integrating differentiable rendering for improved generalization. 




\end{document}